\documentclass[sigconf, nonacm]{acmart}

\usepackage{fancyhdr}
\usepackage{mathtools}
\usepackage{bm}
\usepackage[ruled,noend]{algorithm2e}
\usepackage{caption}
\usepackage{utfsym}
\usepackage{pifont}
\usepackage{enumitem}
\usepackage{subcaption}

\usepackage{tikz}
\newcommand*\circled[1]{\tikz[baseline=(char.base)]{
            \node[shape=circle,fill,inner sep=1.2pt] (char) {\textcolor{white}{#1}};}}

\theoremstyle{plain}
\newtheorem{theorem}{Theorem}[section]
\theoremstyle{definition}
\newtheorem{definition}[theorem]{Definition}
\newenvironment{myProof}{\noindent{\textit{Proof}.}}{\hfill$\square$}

\theoremstyle{corollary}
\newtheorem{corollary}[theorem]{Corollary}

\author {
    Shangyu Liu\textsuperscript{\rm 1},
    Zhenzhe Zheng\textsuperscript{\rm 1},
    Xiaoyao Huang\textsuperscript{\rm 2},
    Fan Wu\textsuperscript{\rm 1},
    Guihai Chen\textsuperscript{\rm 1}
    Jie Wu\textsuperscript{\rm 2}
}
\affiliation {
    \textsuperscript{\rm 1}Shanghai Jiao Tong University\\
    \textsuperscript{\rm 2}Cloud Computing Research Institute, China Telecom\\
    \{liushangyu,zhengzhenzhe\}@sjtu.edu.cn,
    huangxy32@chinatelecom.cn, \\
    \{fwu,gchen\}@cs.sjtu.edu.cn,
    wujie@chinatelecom.cn
    \country{}
}

\begin{document}

\fancyhead{}
\fancyfoot[C]{\thepage}

\title{Efficient Distributed Retrieval-Augmented Generation for Enhancing Language Model Performance}

\begin{abstract}
Small language models (SLMs)
support efficient  deployments on resource-constrained edge devices, but their limited capacity compromises inference performance.
Retrieval-augmented generation (RAG) is a promising solution to enhance model performance by integrating external databases, without requiring intensive on-device model retraining. However, large-scale public databases and user-specific private contextual documents are typically located on the cloud and the device separately, while existing RAG implementations are primarily centralized. To bridge this gap, we propose DRAGON, a distributed RAG framework to enhance on-device SLMs through both general and  personal knowledge without the risk of leaking document privacy. Specifically, DRAGON decomposes multi-document RAG into multiple parallel token generation processes performed independently and locally on the cloud and the device, and employs a newly designed Speculative Aggregation, a dual-side speculative algorithm to avoid frequent output synchronization between the cloud and device. A new  scheduling algorithm is further introduced to identify the optimal aggregation side based on real-time network conditions. Evaluations on real-world hardware testbed demonstrate a significant performance improvement of DRAGON---up to $1.9\times$ greater gains over standalone SLM compared to the centralized RAG, substantial reduction in per-token latency, and negligible Time to First Token (TTFT) overhead.
\end{abstract}
\maketitle
\vspace{-8pt}
\section{Introduction}
Although large language models (LLMs) such as GPT-4~\cite{GPT4} and DeepSeek-V3~\cite{deepseekv3} have demonstrated remarkable performance in real-world applications, their substantial deployment costs have led to predominant cloud-based hosting. As a result, users are required to upload private context along with their queries, raising serious privacy concerns~\cite{forbes}.
Recently, small language models (SLMs) such as Phi-4-mini~\cite{phi4} and Qwen2.5-1.5B~\cite{qwen2}, have emerged as promising alternatives, offering efficient local deployment on edge devices. 
However, although SLMs are notably smaller than cloud-hosted LLMs---leading to reduced performance on both personal and general tasks---they still remain too large for resource-constrained devices to support on-device fine-tuning or training~\cite{kairouz2021advances} to adapt to newly generated data and user feedback.

\begin{figure}[t]
    \centering
    \includegraphics[width=\linewidth]{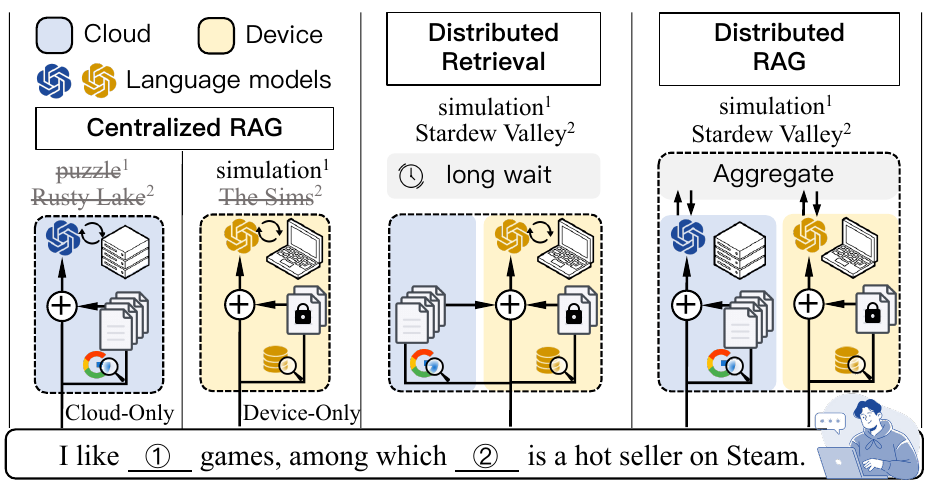}
    \Description{}
    \caption{Comparison between different RAG architectures.}
    \label{fig:scenarios}
    \vspace{-4mm}
\end{figure}

Retrieval-augmented generation (RAG)~\cite{FacebookRAG, ralm} has demonstrated 
effectiveness in boosting the performance of SLMs by incorporating contextually relevant documents from external databases, such as Wikipedia~\cite{wiki_dpr}. The performance gain increases monotonically with the scale of the database~\cite{rag_scaling}, showing an opportunity for SLMs to achieve comparable or even better performance than standalone LLMs~\cite{chen2024benchmarking}. More importantly, by expanding user-specific external database (also known as the non-parametric memory~\cite{FacebookRAG}), model customization and knowledge updates can be achieved efficiently without model training. Typically, large-scale public databases containing general knowledge are hosted 
in the cloud, whereas user-specific private databases are maintained on-device. Since the query context may involve both general and personal data,
it is essential for retrieval-augmented SLMs to support distributed databases located in the  cloud and device. Unfortunately, most existing RAG solutions~\cite{FacebookRAG,ralm,retro} adopt a centralized architecture. Figure~\ref{fig:scenarios} presents an example of game recommendation. 
The cloud-only RAG returns an incorrect game genre, although  private documents indicate a preference for simulation games. In contrast, the device-only RAG fails to retrieve the best-selling game lists without accessing to the general knowledge  in the cloud. 

An intuitive solution, similar to federated search~\cite{federated_search}, is to retrieve documents from the cloud-side database, merge them with those retrieved locally on-device, and perform model  inference in a centralized manner. However, this approach may incur substantial latency overhead considering key-value (KV) caching~\cite{zhao2023survey}, a fundamental mechanism in language model serving that stores intermediate attention states to enable efficient reuse of past computations. Given a context sequence length $n$, the KV cache trades $O(n)$ storage for a reduction in decoding time complexity from $O(n^2)$ to $O(n)$. Therefore, the KVs of documents are often pre-computed, stored in the database, and retrieved. 
This leads to a dilemma: when retrieving the raw text of cloud-side documents, the device must compute their KVs from scratch, incurring significant computation latency; Conversely, directly retrieving the KVs from the cloud introduces substantial transmission latency, as the size of KVs can be comparable to, or even larger than, the model parameters, especially as the number of document grows~\cite{kvquant}.

To address these issues, we propose DRAGON, a \underline{d}istributed \underline{r}etrieval-\underline{a}ugmented \underline{g}enerati\underline{on} framework designed to enhance the performance of on-device language model inference. Following the Law of Total Probability, DRAGON first decomposes the multi-document RAG process into a dual-side workflow,  and then aggregates their output tokens for the final result. In this workflow, the cloud and device  sides independently execute their own LM instances using documents retrieved from their databases. Document KVs are stored and loaded locally without transmission or re-computation, thereby reducing first-token latency and preserving document privacy. Nonetheless, the output aggregation requires frequent exchange of data packets between the cloud and  device at every token generation step, due to the auto-regressive nature of language models. 
This transmission pattern requires a persistent low-latency connection between the cloud and device, which is difficult to guarantee in real-world scenarios~\cite{network_delay, latency_survey}.

To solve this challenge, we draw inspiration from the draft-then-verify paradigm in \textit{Speculative Decoding}~\cite{leviathan2023fast} and propose a new dual-side speculative algorithm, namely \textit{Speculative Aggregation}. In this algorithm, the decoding processes on both sides continuously generates draft tokens, and an Aggregator on either side (depending on certain scheduling criteria) asynchronously verifies and aggregates them. Decoding is interrupted and the corresponding KV states are rolled back for re-computation only when a draft is rejected. As our theoretical analysis proves the equivalence between Speculative Aggregation and the vanilla synchronized version, the end-to-end latency can be reduced by overlapping transmission and decoding processes, especially when the output distributions of the two sides are similar. 

We implement a fully functional distributed RAG workflow and construct a testbed using real-world hardware. Based on this, we evaluate DRAGON against various RAG architectures using representative SLMs on large-scale retrieval corpora and datasets. Experimental results of language modeling on WikiText demonstrate that DRAGON achieves $1.9\times$ and $1.4\times$ greater performance gains over the standalone SLM than the centralized method, using Qwen2.5-1.5B and OPT-1.3B, respectively. Moreover, DRAGON shows strong robustness under various network conditions and achieves a 42.4\%-49.5\% reduction in per-token latency compared to synchronized methods under 300 ms network latency. While retrieving raw text and KV states incurs up to $8.9\times$ and $15.3\times$ overhead in response time, DRAGON introduces negligible overhead. Extensive simulations further verify that the proposed scheduling algorithm achieves increasing delay reduction as network latency grows.

We summarize the key contributions of this work as follows:
\begin{itemize}[leftmargin=*, topsep=2pt]
    \item We propose DRAGON, the first distributed RAG system that 
    supports distributed documents retrieval and collaborative output generation between  cloud and device. It significantly enhances the performance of on-device SLMs with the  integration of both personal and general knowledge.
    \item We introduce \textit{Speculative Aggregation}, a dual-side speculative algorithm that decouples synchronized aggregation from sequential decoding by asynchronously verifying the output alignment between cloud and device, greatly reducing end-to-end latency.
    \item We further design an adaptive scheduling algorithm to dynamically identify the optimal aggregation side under varying network conditions, effectively improving decoding efficiency.
    \item We implement DRAGON in a real-world hardware testbed and perform comprehensive evaluations using representative SLMs and large-scale retrieval corpora, demonstrating significant performance improvements of on-device SLMs with negligible overhead even under high-latency network conditions.

\end{itemize}
\vspace{-3mm}
\section{Preliminaries}
\subsection{Retrieval-Augmented Generation\label{sec:output_aggregation}}
Retrieval-augmented generation~\cite{FacebookRAG} integrates off-the-shelf language models with documents retrieved from an external database to capture long-tail knowledge and keep up-to-date with new information. In traditional LM inference, given an input token sequence $x_{<M}=\{x_0, \dots,x_{M-1}\}$ (indices of tokens in vocabulary $V$) and the maximum context length $N$, the output generation process aims to maximize the probability $\prod_{t=M}^{N-1} p(x_t|x_{<t})$. In order to incorporate external documents, we process each document concatenated with the query separately, and then interpolate the output distributions (termed as \textit{output aggregation}~\cite{FacebookRAG, replug, asai2024reliable})\footnote{The output aggregation is different from \textit{context aggregation}~\cite{ralm}), where external documents are concatenated and prepended to the input query $x_{<t}$ all at once.}. Following the Law of Total Probability, we can derive the interpolation as
\begin{equation}
    \bm{p}(x_t|x_{<t})= \sum\nolimits_{d\sim p(d)}p(d|x_{< t})\cdot \bm{p}(x_t|d,x_{< t}),
    \label{equa:total_prob}
\end{equation}
where 
$p(d|x_{<t})$ denotes the weight of the document $d$ on the output distribution $\bm{p}(x_t|d,x_{< t})$. Since $p(d|x_{<t})$ cannot be directly obtained in practice, we retrieve $d$ from a sufficiently large corpus $\mathcal{D}$ and only consider top-$k$ documents with the highest relevance score $\mathcal{R}_{\mathcal{D}}(d, x_{<t})$. 
Equation~\eqref{equa:total_prob} offers the opportunity to decompose the multi-document RAG workflow into parallel generation processes, 
enabling device-cloud distributed RAG. This decomposition also significantly alleviates the limitation of maximum context length on resource-constraint devices. 

\subsection{Device-Cloud Distributed RAG\label{sec:vanilla_DRAG}}
To enhance the performance of on-device LLM inference, we propose a device-cloud distributed RAG framework based on the above discussed output aggregation paradigm. Given an input $x_{<t}$, we retrieve personalized documents $D^{\text{device}}$ from a device-side private database and then compute the next-token distributions $\bm{P}^{\text{device}}_t = \begin{bmatrix}\bm{p}(x_t|d,x_{<t})\end{bmatrix}^\top_{d\in D^{\text{device}}}$ using an on-device LLM $\mathcal{M}^{\text{device}}$.
In parallel, we employ a similar process in the cloud and obtain the cloud-side next-token distributions $\bm{P}^{\text{cloud}}_t$. After gathering all documents $D=D^{\text{device}}\cup D^{\text{cloud}}$ and their corresponding output distributions $\bm P_t=\begin{bmatrix} \bm P^{\text{device}}_t ,\bm P^{\text{cloud}}_t \end{bmatrix}^\top$, we sample the next token according to
\begin{equation}
    x_t \sim \bm \omega_t^\top \bm P_t=\sum\nolimits_{d\in D} \omega_t(d)\cdot\bm{p}(x_t|d,x_{<t}),
    \label{equa:aggregated_sampling}
\end{equation}
where $\bm \omega_t =\begin{bmatrix}\omega_t(d)\end{bmatrix}^\top_{d\in D}$ denotes the interpolation weights, which are computed based on relevance scores $\mathcal{R}$ as $$\omega_t(d)=\exp \mathcal{R}(d, x_{<t})/\sum\nolimits_{d^\prime \in D}\exp \mathcal{R}(d^\prime, x_{<t}).$$ We refer to this workflow as the vanilla distributed RAG.

Despite its effectiveness, frequent synchronization between the device and the cloud can introduce a substantial latency. On one hand, the tight data coupling in distributed RAG leads to idle waiting, especially when decoding latencies significantly differ due to hardware heterogeneity in cloud and device. 
During the auto-regressive LLM model inference, the output $x_{t-1}$ is expected on both sides as the input for generating $\bm{P}_{t}$.
At each token generation step $t$, computing Equation~\eqref{equa:aggregated_sampling} requires waiting for both the device-side and cloud-side output distributions, $\bm P_t^{\text{device}}$ and $\bm P_t^{\text{cloud}}$.
On the other hand, frequent transmission of  data packets makes this device-cloud distributed RAG paradigm highly sensitive to network stability. Transmitted data packets at each step includes a 2-byte integer representing the token $x_t$ and a float matrix $\bm P_t$ encoding the output distributions\footnote{The float matrix $\bm P_t$ has a size of $|V|\max(|D^{\text{device}}|, |D^{\text{cloud}}|)$, where the vocabulary size $|V|$ is typically less than 50,000.}. Due to the small data packet size, transmission time is often dominated by data-independent factors~\cite{RTT, cardwell2016bbr}, such as the connection round-trip time (RTT). Finally, idle waiting and transmission latency at each token generation step accumulate over a long output sequence, significantly amplifying the overall overhead.

\subsection{Problem Formulation}

We define the LLM inference as a distributed process where the device-side and cloud-side token generation processes, $\mathcal{F}^{\text{device}}$ and $\mathcal{F}^{\text{cloud}}$, executes alternatively. We assume the final output token sequence is generated on-device by sampling $x\sim \bm p_t$.
Let $A_t$ be an auxiliary set for transferring information between the device and the cloud at iteration $t$, which is initially empty, and let $\bm{p}_t$ denote the next-token distribution. The workflow can then be expressed as $A^{\text{device}}_t, \bm p_t \leftarrow \mathcal{F}^{\text{device}}(A^{\text{cloud}}_{t-1}, \mathcal{M}^{\text{device}}, D^{\text{device}}, x_{<t})$ on the device, and $A^{\text{cloud}}_t \leftarrow \mathcal{F}^{\text{cloud}}(A^{\text{device}}_t, \mathcal{M}^{\text{cloud}}, D^{\text{cloud}}, x_{<t})$ on the cloud, respectively. Finally, the optimization objective is given by
\begin{equation}
    \min_{\mathcal{F}} \frac{1}{N}\sum\nolimits_{t=1}^N \left(-p(x^*_t|x_{<t})\log \bm p_t(x^*_t|x_{<t}) + \lambda C(A_t,\mathcal{F})\right),
    \label{equa:objective}
\end{equation}
where $x^*_t$ represents the optimal token at step $t$ and $C$ denotes the end-to-end latency per token resulted from the transmission of $A_t$ and execution of $\mathcal{F}$ between the cloud and device. The coefficient $\lambda$ controls the trade-off between performance and efficiency. 

\section{Overview of DRAGON\label{sec:overview}}
\begin{figure}[t]
    \centering
    \includegraphics[width=\linewidth]{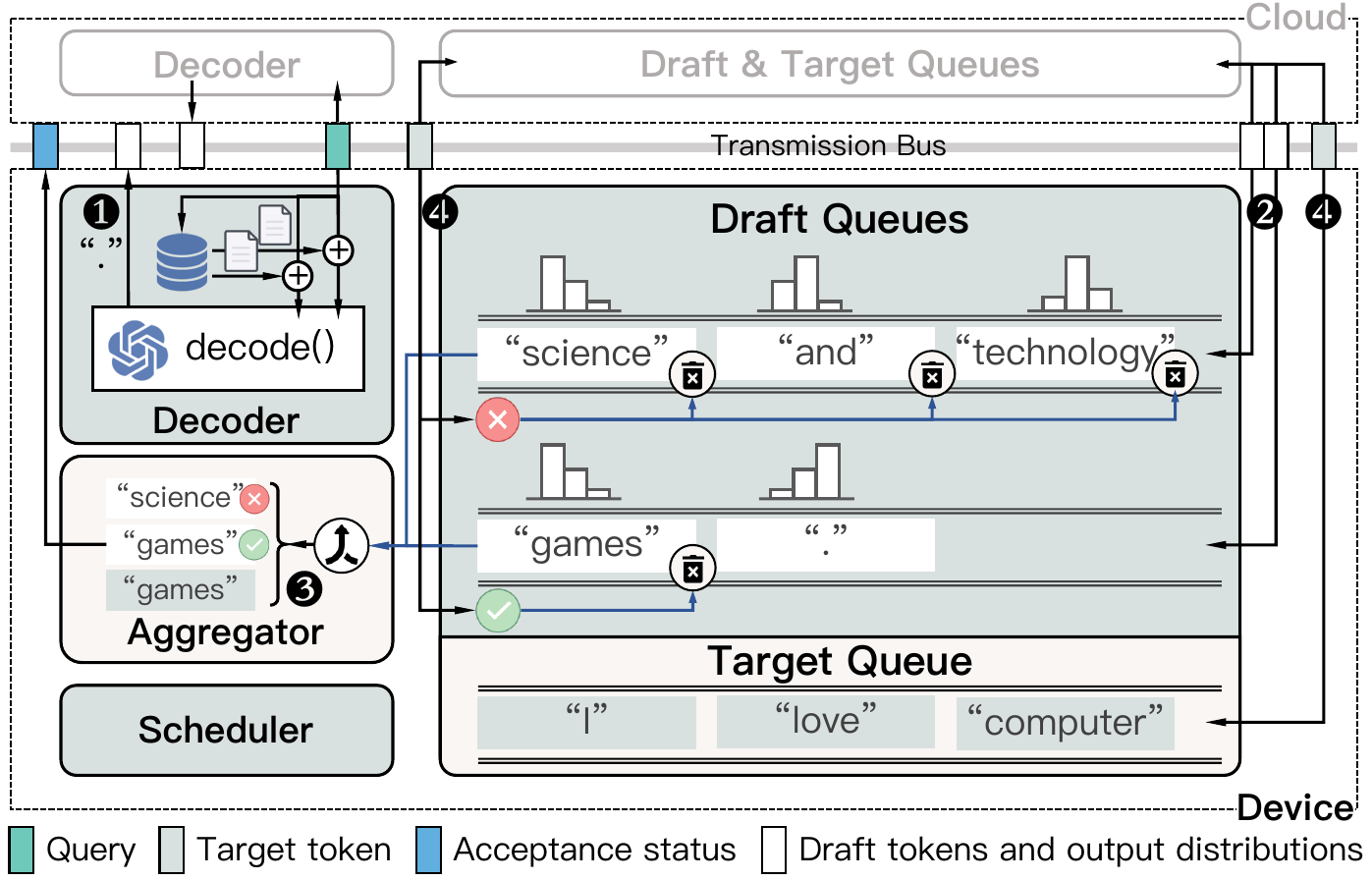}
    \Description{}
    \vspace{-5mm}
    \caption{Overview of the DRAGON framework.}
    \vspace{-5mm}
    \label{fig:overview}
\end{figure}

To enhance on-device LLM inference while minimizing the latency overhead, we propose DRAGON, a device-cloud distributed RAG framework. In this framework, we sample tokens from distributions aggregated from the device-side and cloud-side RAG outputs, enabling an  integration of personalized information and generic knowledge.
To mitigate the inherent latency caused by frequent device-cloud synchronizations in vanilla distributed RAG, 
we perform distribution aggregation and next-token sampling in a speculative manner, where draft tokens are generated on both sides and then verified on either side. 
Accordingly, as shown in Figure~\ref{fig:overview}, DRAGON consists of four modules deployed on both sides, including Decoders, Queues, Profilers, and Schedulers, along with a device/cloud-switchable Aggregator on either side. 

We organize  Decoders, Queues and  Aggregator by a producer-consumer paradigm, enabling asynchronous decoding of draft tokens. The Decoder serves as a token producer, and on each side $s\in\{\text{device}, \text{cloud}\}$ it decodes draft tokens $x^s_t$ independently based on locally-aggregated output distributions $\bm p^s_t=(\tilde{\bm\omega}_t^s)^\top \bm P^s_t$ where $\tilde{\bm\omega}_t=\begin{bmatrix}\omega_t(d)\end{bmatrix}^\top_{d\in D^s}$, similar to Equation~\eqref{equa:aggregated_sampling} but using the  retrieved local documents $D^s$ only (\circled{1}). The draft tokens $x^s_t$ and their corresponding distribution vectors $\bm p^s_t$ are broadcast to the other side.
On each side, we enqueue $x^{s}_t$ 
into Draft Queues (\circled{2}). 
The Aggregator, as a consumer, continuously consumes draft tokens from the front of local queues and performs aggregation process
(\circled{3}). Subsequently, the aggregation results of the draft token are broadcast to Draft Queues on both sides. For each queue, the first token is dequeued if accepted, or the entire queue is cleared if rejected. The final target token output by Aggregator is 
enqueued into Target Queue on both sides (\circled{4}). Although the dependencies between the aggregator and decoder cannot be eliminated, the data transmission latency can be overlapped with the decoding time, mitigating idle waiting.
To accommodate dynamic computing resources on both sides and  network bandwidth between them, we further design Profilers and Schedulers to  identify the optimal aggregation side. 

\section{Speculative Aggregation\label{sec:speculative_aggregation}}
Inspired by \textit{Speculative Decoding}~\cite{leviathan2023fast}, we propose \textit{Speculative Aggregation} to reduce the device-cloud communication latency. \textit{Speculative Decoding} adopts a draft-then-verify decoding paradigm to reduce the number of calls to the resource-intensive LLM. Similarly, \textit{Speculative Aggregation} utilizes two independent decoding processes, the device-side and cloud-side Decoders, to draft multiple candidate future tokens, which are then verifies through an Aggregator. This is equivalent to directly sampling from the distributions aggregated from the device-side and cloud-side outputs. As the aggregation involves collecting output distributions over the network, we expect the speculative algorithm to reduce its frequency and mitigate data transmission costs.

More specifically, the Aggregator stays in a blocked wait state until both local Draft Queues are non-empty. Once this condition is met, it retrieves one token $x_t^{\text{device}}$/$x_t^{\text{cloud}}$ from the front of each queue and fetches corresponding locally-aggregated output distributions $\bm p_t^{\text{device}}$/$\bm p_t^{\text{cloud}}$ from the cache. The tokens and the distributions are then provided as inputs to the aggregation. Subsection~\ref{sec:target_distribution} presents the target distribution of the aggregation while Subsection~\ref{sec:strategy} introduces an speculative strategy to sample from the target distribution equivalently. Subsection~\ref{sec:ac_rate} analyzes the acceptance probability, providing guidance for further scheduling design. Since the workflows of the device and cloud sides are designed to be symmetric, we define $\{l, r\} = \{\text{device}, \text{cloud}\}$ to maintain generality and avoid repetition. From the perspective of the Aggregator, $l$ refers to the local side that performs aggregation, while $r$ denotes the remote side, which only generates draft tokens.

\subsection{Target Token Distribution\label{sec:target_distribution}}
The objective of speculative aggregation is to generate tokens that are equivalent to those sampled from the target distribution $\bm{p}_t=\bm\omega_t^\top\bm P_t$ as defined in Equation~\eqref{equa:aggregated_sampling}. We partition $\bm P_t$ block-wise, grouping its distribution vectors by generation side, and have 
$\bm p_t= (\bm\omega^l_t)^\top \bm P^l_t+(\bm\omega^r_t)^\top \bm P^r_t$.
For each $s\in\{l,r\}$, we have $\bm\omega_t^l=\eta^s_t\tilde{\bm\omega}_t^l$ where $\eta^s_t=h^s_t/(h^l_t+h^r_t)$ and $h^s_t=\sum_{d\in D^s} \exp \mathcal R(d, x_{<t})$. As a result, given the locally-aggregated output distributions $\bm p_t^{l}$ and $\bm p_t^{r}$ (\S~\ref{sec:overview}), the target distribution $\bm p_t$ can be obtained by an interpolation:
\begin{equation}
    \bm p_t=\eta^l_t\bm p^l_t+\eta^r_t\bm p^r_t.
    \label{equa:target_distribution}
\end{equation} 
To align with this computation process, on each side $s\in\{l,r\}$, a corrected value\footnote{We adopt the log-sum-exp trick to maintain numerical stability (See Appendix~\ref{sec:numerical_stable}).} of ${h}^s_t$ is computed and retained during decoding $x^s_t$, and then broadcast and stored along with draft tokens and the locally-aggregated distributions. 

\noindent\textbf{Dynamic weights.} The interpolation weights $\eta^l_t$ and $\eta^r_t$ in Equation~\eqref{equa:target_distribution} can be dynamic, as the relevance between documents and the ongoing context may vary during generation. While current studies~\cite{FacebookRAG, replug} that employ output aggregation adopt static document relevance scores, we explore dynamic weights by adopting a strategy inspired by those used in recommendation systems. Upon receiving the input query $x_{<M}$, each side $s\in \{l, r\}$ performs a one-time retrieval of a relatively large document set $D^s$ (as in most existing works~\cite{lewis2020retrieval,izacard2020leveraging,replug}) to avoid key-value recomputation caused by changes in the document prefix. During the decoding of draft token $x_t$, we re-estimate the relevance scores $\mathcal{R}(d, x_{<t})$ for $d\in D^s$ using a local re-ranking model (e.g., a Cross-Encoder~\cite{sentence-transformer}) and re-calculate the corrected $h^s_t$ before transmission.

\subsection{Design of Aggregation Strategy\label{sec:strategy}}
To sample $x_t\sim\bm p_t$, we instead perform two independent speculative sampling processes as follows:
\begin{itemize}[leftmargin=*]
    \item Keep the draft token $x^l_t$ as $ \tilde x^l_t $ if $ \bm p_t^l({x}^l_t) \leq \bm p_t^r({x}^l_t) $, and in case $\bm p_t^l({x}^l_t) > \bm p_t^r({x}^l_t)$ we reject the sample with probability $ \eta^r_t ( 1 - \bm p_t^r({x}^l_t)/\bm p_t^l({x}^l_t) ) $ and sample $\tilde x_t^l$ again from an adjusted distribution $ \tilde{\bm p}_t^l=\text{norm} ( \max(0, \bm p_t^r - \bm p_t^l) ) $.
    
    \item Keep the draft token $x^r_t$ as $ \tilde x^r_t $ if $ \bm p_t^r({x}^r_t) \leq \bm p_t^l({x}^r_t) $, and in case $\bm p_t^r({x}^r_t) > \bm p_t^l({x}^r_t)$ we reject the sample with probability $ \eta^l_t ( 1 - \bm p_t^l({x}^r_t)/\bm p_t^r({x}^r_t) ) $ and sample $\tilde x_t^r$ again from an adjusted distribution $ \tilde{\bm p}_t^r=\text{norm}(\max(0, \bm p_t^l - \bm p_t^r))$.
\end{itemize}
It is straightforward to show\footnote{The proof is included in Appendix~\ref{sec:proof_1}.} that through these sampling processes, both $ \tilde x^l_t $ and $ \tilde x^r_t $ are indeed drawn from the aggregated distribution $\eta^l_t\bm p^l_t+\eta^r_t\bm p^r_t$. We select either $\tilde x^l_t$ or $\tilde x^r_t$ as $x_t$ with uniform probability, ensuring $x_t\sim \bm{p}_t$. Finally, each draft token $x^l_t$ and $x^r_t$ is accepted if it matches the target token $x_t$; otherwise, it is rejected. The aggregation strategy at each step $t$ is summarized in Algorithm~\ref{alg:aggregation}. It is worth noting that we design a sampling-based method rather than simply selecting between $x^l_t$ and $x^r_t$, in order to ensure that $x_t\sim \bm p_t$ holds. A counterexample for binary selection is illustrated in cases where $\arg\max {\bm p_t}$ differs from both $\arg\max{\bm p^l_t}$ and $\arg\max{\bm p^r_t}$.

\begin{algorithm}[t!]
    \caption{SpeculativeAggregation}
    \label{alg:aggregation}
    \KwIn{Draft tokens $x_t^s$, locally-aggregated distributions $\bm p_t^s$, and aggregation weights $h^s_t$, for $s\in\{l, r\}$}
    \KwOut{Target token $x_t$, acceptance status $\mathcal S^l$ and $\mathcal S^r$}
    \SetKwProg{Fn}{Function}{:}{}

    \BlankLine
    \SetKwFunction{Sample}{Sample}
    \Fn{\Sample{$x$, $\bm p^a$, $\bm p^b$, $\eta$}}{
        $\tilde x \leftarrow x$, $\sigma^a\sim U(0, 1)$\;
        \If{$ \bm p^a(x) > \bm p^b(x) $, $\sigma^a< \eta ( 1 - \frac{\bm p^b(x)}{\bm p^a(x)}) $}{
            $\tilde x\sim \text{norm} ( \max(0, \bm p^b - \bm p^a) ) $\;
        }
        \KwRet $\tilde x$\;
    }
    
    \BlankLine
    $\eta^l_t\leftarrow h^l_t/(h^l_t+h^r_t)$, $\eta^r_t\leftarrow 1-\eta^l_t$\;
    $\tilde x_t^l\leftarrow \text{Sample}(x^l_t, \bm p^l_t, \bm p^r_t, \eta^r_t)$, $\tilde x_t^r\leftarrow \text{Sample}(x^r_t, \bm p^r_t, \bm p^l_t, \eta^l_t)$\;
    $\sigma \sim U(0, 1)$, $x_t\leftarrow \tilde x_t^l\cdot \bm 1_{\sigma\leq 0.5}+\tilde x_t^r\cdot\bm{1}_{\sigma>0.5}$\;
    $\mathcal S^l\leftarrow x_t^l=x_t$, $\mathcal S^r\leftarrow x_t^r=x_t$\;
    \KwRet $x_t$, $\mathcal{S}^l$, $\mathcal{S}^r$\;
\end{algorithm}

We now present a general procedure for sampling multiple consecutive tokens. At each step $t$, the following workflow is executed:
\begin{enumerate}[label=\arabic*),leftmargin=*,topsep=2pt]
    \item \label{workflow:start}The Aggregator waits until both Draft Queues are non-empty, then fetches $x^s_t$ from the front of the local Draft Queues and retrieves the auxiliary variables $\bm p^s_t$ and $h^s_t$ from the local cache, for each $s\in\{l,r\}$.
    \item The Aggregator performs aggregation as defined in Algorithm~\ref{alg:aggregation}. The outputs, including the target token $x_t$ and the acceptance status of each draft token, are broadcast to notify both sides.
    \item Upon receiving the message, each side checks the acceptance status of both $x^l_t$ and $x^r_t$. If a token is accepted, it is dequeued from the corresponding Draft Queue and step~\ref{workflow:update_t} is executed; otherwise, step~\ref{workflow:reject} is executed.
    \item \label{workflow:reject}If $x^s_t$ is rejected, its corresponding Draft Queues on both sides are cleared and the side $s$ rolls back its KV cache and re-computes the next draft token $x^s_{t+1}$ using the target token $x_t$ as input.
    \item \label{workflow:update_t}Update step $t\leftarrow t+1$, and go back to step~\ref{workflow:start}.
\end{enumerate}

We adopt a pipeline approach rather than performing aggregation in parallel. In centralized \textit{Speculative Decoding}, each execution of the target LLM requires waiting for the draft model to generate the current token and for the LLM to verify the previously generated one. By verifying consecutive tokens in parallel, multiple LLM inferences can be merged into a single pass, shifting the primary latency bottleneck from target LLM inference to the sequential decoding of draft tokens. Conversely, in \textit{Speculative Aggregation} for distributed RAG, the time to the next token is dominated by data transmission over the network. Consecutive transmission of small data can be naturally overlapped since each transmission does not significantly occupy the I/O for an extended period. Parallelizing the aggregation process instead introduces waiting between draft tokens until the batch is fully populated. We employ queues to construct a pipeline, where each draft token is transmitted and enqueued immediately upon generation, ensuring it is verified at the earliest opportunity.

\subsection{Analysis of Acceptance Rate\label{sec:ac_rate}}
We now analyze the factors that influence the acceptance rate of draft tokens on both the device and the cloud sides. 

\begin{definition}
For $s\in\{l,r\}$, the acceptance rate $ \beta^s_t $, is the probability of accepting $ x^s_t \sim \bm p^s_t=\sum_{d\in D^s} \omega_t(d)\bm p(x_t|d,x_{<t})$ by the aggregation strategy, given a prefix $ x_{<t} $. 
\end{definition}

First, we consider the side $l$ as an example. The acceptance of the draft token $x^l_t$, sampled from $\bm p^l_t$ by the Decoder, can be classified into two cases: i) it is accepted during the speculative sampling of $\tilde{x}^l_t$ and ii) it is output by the speculative sampling of $\tilde{x}^r_t$, where either $x^r_t=x^l_t$ and is accepted, or $x^l_t\sim \tilde{\bm p}^r_t$. Let $\gamma^l$ and $\gamma^r$ represent the weights assigned to $\tilde{x}^l_t$ and $\tilde{x}^r_t$ in the random selection following these sampling processes ($\gamma^l+\gamma^r=1$). We adopt the definition of divergence from \cite{leviathan2023fast}, given by $ \delta = D_{LK}(\bm p^l_t, \bm p^r_t) = 1 - \sum_x \min(\bm p^l_t(x), \bm p^r_t(x)) $. The expected acceptance rate $\alpha^l_t=\mathbb E_{x\sim \bm p^l_t(x)}(\beta^l_t)$ is computed as
\begin{equation}
     \alpha^l_t= \gamma^l (1-\eta^r_t\delta)+\gamma^r\sum\nolimits_x \bm p^l_t(x)\bm p_t(x),
    \label{equa:acceptance_rate}
\end{equation}
where the two terms represent the acceptance probability of the two cases above, respectively. These terms are mutually-exclusive and influenced by the mixture weights $\gamma^l$ and $\gamma^r$.

\begin{theorem}
    Given any distributions $\bm p^l_t$ and $\bm p^r_t$, when $\eta^r_t$ is fixed, maximizing $\alpha^l_t$ is equivalent to maximizing $\gamma^l$.
    \label{theorem:gamma}
\end{theorem}

\begin{myProof}
Substituting $\bm p_t=\eta^l_t\bm p^l_t+\eta^r_t\bm p^r_t$\footnote{For brevity, the variable $x$ is omitted in the distribution notation throughout the following proof.} and subtracting the two terms in Equation~\eqref{equa:acceptance_rate} yields $(1-\eta^r_t\delta)-\sum \bm p^l_t\bm p_t=\eta^l_t(1-\sum\bm (p_t^l)^2)+\eta^r_t\sum(\min(\bm p_t^l,\bm p_t^r)-\bm p_t^l\bm p^r_t)$. Since $1>\sum_x(\bm p_t^l)^2$, $\min(\bm p^l_t,\bm p^r_t)\geq \bm p^l_t\bm p^r_t$ and $0\leq\eta^l_t,\eta^r_t\leq1$, it follows that $1-\eta^r_t\delta\geq\sum\bm p^l_t\bm p_t$ always holds, with equality holding only when $\eta^l_t=0$, $\eta^r_t=1$ and $\delta=0$. This condition implies that the two distributions $\bm p^l_t$ and $\bm p^r_t$ are completely disjoint. Consequently, maximizing $\gamma^l$ leads to the maximization of the expected acceptance rate $\alpha^l_t$.
\end{myProof}

For the side $r$, Theorem~\ref{theorem:gamma} holds symmetrically, where maximizing the acceptance of $x^r_t$ corresponds to maximizing $\gamma^r$. Clearly, the objectives on sides $l$ and $r$ conflict with each other. For simplicity in framework design, we adopt $\gamma^l = \gamma^r = 0.5$ to strike a balance (as shown in Algorithm~\ref{alg:aggregation}).

The expected acceptance rate is then influenced by the degree of overlap between the draft distributions on the two sides. When the distributions $\bm p^l_t$ and $\bm p^r_t$ perfectly coincide, i.e., the divergence $\delta$ becomes zero, the first term of Equation~\eqref{equa:acceptance_rate}, $1-\eta^r_t\delta$, reaches its maximum value. Simultaneously, since the second term follows
\begin{equation*}
\sum\nolimits_x\bm p^l_t(x)\bm p_t(x)\leq \sqrt{\sum\nolimits_x\bm p^l_t(x)^2\sum\nolimits_x\bm p_t(x)^2},
\end{equation*}
based on the Cauchy-Schwarz inequality and achieves its maximum when $\bm p^l_t(x)=\bm p^r_t(x)=\bm p_t(x)$, the expected acceptance rate is maximized.
Conversely, when the support sets of the two distributions are completely disjoint, i.e., $\delta=1$, the product $\bm p^l_t(x)\bm p_t(x)$ becomes zero for every $x$, resulting in a minimized expected acceptance rate. 

This characteristic provides insight into the principle behind \textit{Speculative Aggregation}: we assume that the device-side and cloud-side RAG workflows generate similar results by default, allowing them to asynchronously decode the next tokens without aggregation. Only when they disagree with each other, the acceptance is adjusted by their aggregation weights $\eta^l_t$ and $\eta^r_t$.

\section{Greedy Scheduling\label{sec:scheduling}}
To further minimize the latency $C(A_t,\mathcal{F})$ in Equation~\eqref{equa:objective}, We adaptively schedule which side performs the next aggregation after the current one is completed. The principle behind this is to maximize the overlap between the device-side and cloud-side decoding and transmission processes, jointly considering dynamic computing resources, network bandwidth, and acceptance of draft tokens.

\begin{table}[t]
  \begin{tabular*}{\linewidth}{p{4em}p{4em}l}
    \toprule
    \textbf{$x^l_{t-1}$} & \textbf{$x^r_{t-1}$} & \textbf{Waiting Time for $x^l_t$ and $x^r_t$} \\
    \midrule
    rejected & accepted & $\max(c^l_{\text{dec}},\varphi(c^r_\text{dec}+c^r_\text{trans}))$ \\
    accepted & rejected & $\max (\varphi(c^l_\text{dec}),c^l_\text{trans}+c^r_\text{dec}+c^r_\text{trans})$ \\
    accepted & accepted & $\max (\varphi(c^l_\text{dec}), \varphi(c^r_\text{dec}+c^r_\text{trans}))$ \\
    rejected & rejected & $\max(c^l_\text{dec}, c^l_\text{trans}+c^r_\text{trans}+c^r_\text{dec})$ \\
    \bottomrule
  \end{tabular*}
  \caption{Waiting time for the next pair of draft tokens $x^l_t$ and $x^r_t$ under different acceptance scenarios of the previous draft tokens $x^l_{t-1}$ and $x^r_{t-1}$.}
  \label{tab:ttnt}
  \vspace{-2em}
\end{table}

\subsection{Scheduling Strategy\label{sec:scheduling_strategy}}
Since predicting future acceptance is challenging due to dynamic document relevance and LLM outputs, we employ a greedy strategy, where at each step, we minimize the expected latency per token based on current observations.

The latency per token, denoted as $Z_t$, is computed as the average duration between two consecutive aggregations. It can be viewed as the waiting time for the next pair of draft tokens, $x_t^\text{device}$ and $x_t^\text{cloud}$, including both decoding and transmission delays, as the aggregation duration is negligible. For each side $s\in \{\text{device}, \text{cloud}\}$, let $c^{s}_{\text{dec}}$ denote the decoding delay of a draft token $x^s_t$, and $c^s_{\text{trans}}$ denote the transmission delay of this token and its auxiliary variables from $s$ to the other side. Since the decoding and transmission processes are asynchronous, they may still be ongoing when the scheduling algorithm is executed. Therefore, we define $\varphi(T_\text{total}(u))=\max(0,T_\text{total}(u)+T_\text{begin}(u)-T_\text{now})$ as a function that estimates the remaining time of the total duration $T_\text{total}$ to complete the process $u$, where $T_\text{begin}$ and $T_\text{now}$ are the beginning and current timestamps, respectively. Let $l$ be the side that currently performs aggregation and $r$ be the other one. The best side is then selected as
\begin{equation}
    s^*=\arg \min\nolimits_{s\in \{l,r\}} Z^s_t(\varphi,c^l_{dec},c^l_{trans},c^r_{dec},c^r_{trans}),
    \label{equa:scheduling}
\end{equation}
where $Z^s_t$ denotes the latency per token when $s$ continuously performs the aggregations in the future.

Next, we present the calculation of $Z^s_t$. Table~\ref{tab:ttnt} illustrates the waiting time for the next pair of draft tokens after a previous aggregation. To estimate an averaged $Z^s_t$ over multiple future steps, rather than enumerating all possible combinations of acceptance scenarios, we assume each acceptance scenario repeats continuously\footnote{Please refer to Appendix~\ref{sec:pipeline} for pipeline illustrations of different cases.} and occurs with an expected probability given by the acceptance rate. Therefore, the waiting time in Table~\ref{tab:ttnt} can be simplified to eliminate the function $\varphi$. First, assuming that draft tokens from $r$ are always accepted, the decoding process for consecutive draft tokens will be continuous on $r$. In other words, the decoding of $x^r_t$ begins exactly when $x^r_{t-1}$ is decoded and ready for transmission. Therefore, we have $\varphi(c^r_\text{dec}+c^r_\text{trans})=(T_\text{begin}+c^r_\text{trans}-T_\text{now})+c^r_\text{dec}=c^r_\text{dec}$. Moreover, since the aggregation process can exhaustively consume the token pairs in the Draft Queues, $\varphi(c^l_\text{dec})<c^l_\text{dec}$ holds only when the waiting time for $x^r_t$ dominates. Hence, $\max(\varphi(c^l_\text{dec}), \cdot)=\max(c^l_\text{dec},\cdot)$. Finally, $Z^l_t$ is calculated as
\begin{equation}
    \alpha^r_t\max(c^l_\text{dec},c^r_\text{dec})+(1-\alpha^r_t)\max(c^l_\text{dec}, c^r_\text{dec}+c^l_\text{trans}+c^l_\text{trans}).
    \label{equa:ttnt}
\end{equation}
Symmetrically, $Z^r_t$ is computed by exchanging $l$ and $r$ in Equation~\eqref{equa:ttnt}. Based on this, we can conclude that when the local decoding latency $c^l_\text{dec}$ cannot cover the waiting time for draft tokens from the other side, i.e., $c^l_\text{dec}<c^r_\text{dec}+c^l_\text{trans}+c^l_\text{trans}$, minimizing the overall latency $Z^l_t$ requires maximizing the acceptance rate $\alpha^r_t$.

\begin{figure}[t]
    \centering
    \includegraphics[width=\linewidth]{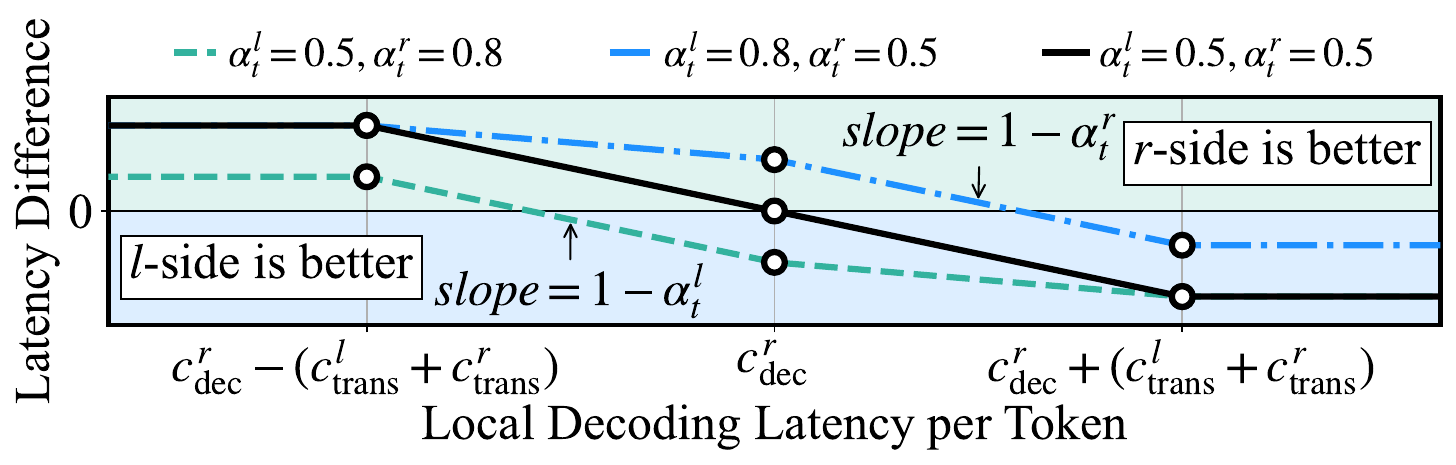}
    \Description{}
    \vspace{-7mm}
    \caption{Difference in per-token latencies when side $l$ and $r$ performs aggregation, versus varying $l$-side decoding latency.}
    \label{fig:latency_diff}
    \vspace{-5mm}
\end{figure}

To decide the optimal side in Equation~\eqref{equa:scheduling}, we calculate the difference in latencies per token when side $l$ and $r$ performs aggregation. The result is presented as a piecewise function,
\begin{equation}
    \Delta Z_t=\begin{cases} 
(1-\alpha^r_t)\text{rtt}, &c^l_\text{dec}\leq c^r_\text{dec} -\text{rtt} \\
(1-\alpha^l_t)j+(\alpha^l_t-\alpha^r_t)\text{rtt}, &c^r_\text{dec} -\text{rtt} < c^l_\text{dec}\leq c^r_\text{dec} \\
(1-\alpha^r_t)j+(\alpha^l_t-\alpha^r_t)\text{rtt}, &c^r_\text{dec}< c^l_\text{dec} \leq c^r_\text{dec} + \text{rtt} \\
(a^l_t-1)\text{rtt}, &c^r_\text{dec}+\text{rtt}<c^l_\text{dec}
\end{cases},
\label{equa:delta_z}
\end{equation}
where $\text{rtt}=c^l_\text{trans}+c^r_\text{trans}$, and $j$ is the difference in decoding latencies, $c^r_\text{dec} - c^l_\text{dec}$. Accordingly, we select side $r$ for aggregation when $\Delta Z_t>0$, and side $l$ otherwise.
Figure~\ref{fig:latency_diff} shows the influence of varying acceptance rates on $\Delta Z_t$. As the acceptance rate of draft tokens from one side increases, the Scheduler tends to favor the opposite side. Moreover, the relationship between $c^l_\text{dec}$ and $c^r_\text{dec}$ also influences the strategy. For instance, when the decoding process on one side becomes the latency bottleneck, aggregation is always performed on that side, which is demonstrated by $(1-\alpha^r_t)\text{rtt}\geq0$ and $(\alpha^l_t - 1)\text{rtt}\leq0$. Clearly, our strategy minimizes the likelihood of repeated bottleneck decoding due to rejection, while maximizing the overlap between the decoding and transmission processes across the two sides.

\subsection{Profiling}

The Profiler helps estimate a set of parameters required to compute Equation~\eqref{equa:scheduling}, including the decoding delay ($c_\text{dec}$) on both the device and cloud sides, the transmission delay ($c^\text{device}_\text{trans}+c^\text{cloud}_\text{trans}$) between them, and the acceptance rates ($\alpha$). The Profiler operates in two stages: i) offline and ii) runtime.

\noindent\textbf{Offline stage.} For each side $s\in \{\text{device}, \text{cloud}\}$, the Profiler measures the decoding delay by simulating the output-aggregation RAG workflow locally. 
We randomly generate $|D^s|$ dummy text chunks as retrieved documents with the same chunk size $M$ as in the real corpus. We use the dummy text directly as the prefix (without a query prompt) and prefill its KV cache in advance. Next, we perform auto-regressive decoding with the same batch size as during runtime, until the context length reaches its maximum $N$. We record the decoding delay $c^s_\text{dec}(t)$ at each step $t=M,\dots,N-1$ by averaging over multiple runs and fit the records $\bm c^s_\text{dec}=\begin{bmatrix}c^s_\text{dec}(t)\end{bmatrix}_{t\in \bm t}$ using a least square errors (LSE) estimator, $\hat{c}^s_\text{dec}=k^s_at/k^s_b+k^s_c$, where the coefficients $k^s_a$, $k^s_b$, and $k^s_c$ are synchronized across both sides. 

We model the transmission delay as $\hat{c}^s_\text{trans}(g)=L^s+g/B^s$, where $g$ represents the data size and $L^s$ and $B^s$ correspond to the network latency and bandwidth for transmitting data from $s$ to the other side. Since one-way delay is hardly measurable, we measure the bi-directional delay $\hat c^\text{device}_\text{trans}+\hat c^\text{cloud}_\text{trans}$ altogether. We utilize \textit{sockperf}
to evaluate the round trip time $L^\text{device}+L^\text{cloud}$ and use \textit{iperf3} to measure the bi-directional bandwidths.

\noindent\textbf{Runtime stage.} To assess decoding latency at runtime, the Decoder on each side $s\in \{\text{device},\text{cloud}\}$ measures the duration $\tilde{c}^s_\text{dec}(t)$ of decoding a draft token at step $t$ using the \textit{time.perf\_counter} function in Python. This measurement is then piggybacked onto the draft token message for convenient synchronization. 
Next, the value of $\hat{c}^s_\text{dec}$ is re-estimated with the intercept $k^s_c$ frozen and the slope updated as
$$\frac{(1-\zeta)k^s_a + \zeta(\tilde{c}^s_\text{dec}(t) - k^s_c)t}{(1-\zeta)k^s_b+\zeta t^2},$$ where $\zeta$ is the weight on new observation. The estimation of transmission delay is refined by means of two moving averages: a real-time estimate and a historical moving average. For the former, we update the round-trip time measurement $L^\text{device}+L^\text{cloud}$ at each step $t$ using the ICMP-based
network diagnostic tool, \textit{ping}. In contrast, the sending and receiving bandwidth $B^{s}$ are updated using \textit{iperf3} every few tokens to avoid excessive overhead. Similarly, we estimate acceptance rates using Equation~\ref{equa:acceptance_rate} and apply a moving average to prevent abrupt changes.

\section{Theoretical Wall-Time Improvement}\label{sec:improvement}
In this section, we present a theoretical analysis to demonstrate the improvement in wall-time efficiency achieved by DRAGON over the vanilla distributed RAG framework described in \S~\ref{sec:vanilla_DRAG}. Specifically, the synchronized aggregation strategy used in the vanilla RAG can be viewed as a special case of speculative aggregation in which draft tokens from both sides are consistently rejected. To facilitate analysis, we assume the aggregation is always performed on the device in following discussions.

\begin{definition}
    Let $Z_t$ and $\tilde{Z}_t$ be the expected per-token latencies at step $t$ when using DRAGON and the vanilla distributed RAG, respectively. Define the speedup as $S_t=\tilde{Z}_t/Z_t$.
\end{definition}

\begin{theorem}
    Given $l=\text{device}$ and $r=\text{cloud}$, the speedup can be described as a piecewise function dependent on the relationship among $c^l_\text{dec}$, $c^r_\text{dec}$ and $\text{rtt}$, as follows:
    \begin{equation}
        \frac{1}{S_t}=\begin{cases} 
        1-\frac{\alpha^r_t}{1+c^r_\text{dec}/\text{rtt}}, &c^l_\text{dec}\leq c^r_\text{dec}\\
        1-(1-\frac{c^l_\text{dec}}{c^r_{dec}+\text{rtt}})\alpha^r_t, &c^r_\text{dec} < c^l_\text{dec} \leq c^r_\text{dec} + \text{rtt}\\
        1, &c^r_\text{dec}+\text{rtt} < c^l_\text{dec}\\
        \end{cases}
    \end{equation}
    \label{theorem:improvement}
\end{theorem}

\begin{myProof}
$Z_t$ is computed according to Equation~\eqref{equa:ttnt}. By substituting $\alpha^l_t=\alpha^r_t=0$ and we obtain $\tilde{Z}_t=\max(c^l_\text{dec},c^r_\text{dec} + \text{rtt})$. The result then follows from a simple case-by-case analysis.
\end{myProof}

\begin{figure}[t!]
    \centering
    \includegraphics[width=\linewidth]{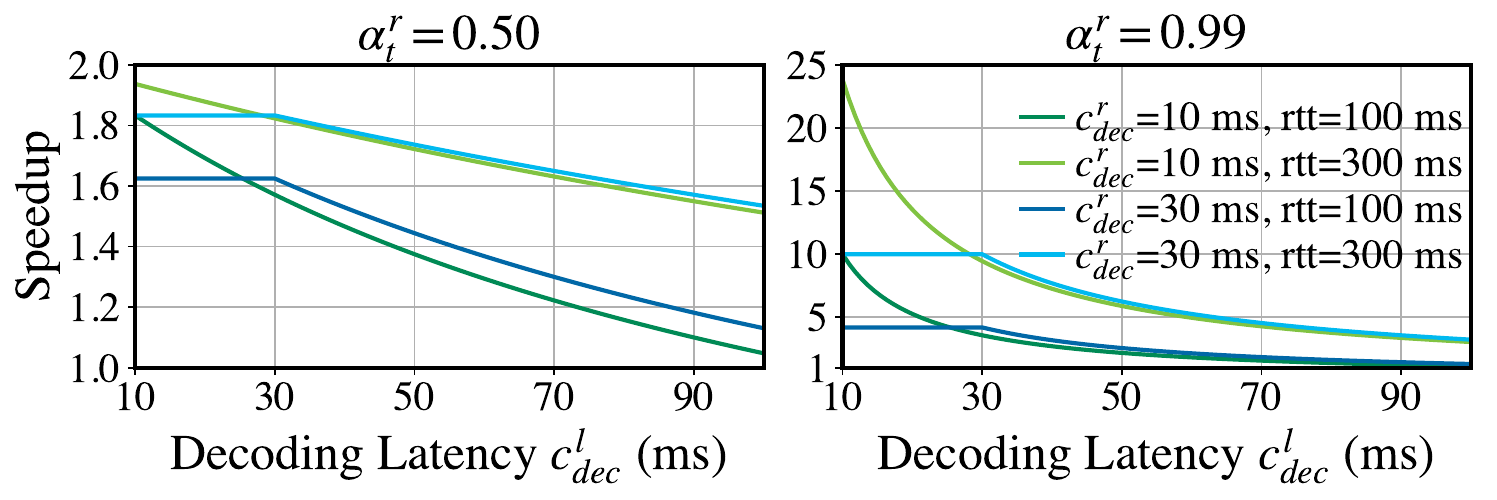}
    \vspace{-0.1in}
    \Description{}
    \caption{Theoretical speedup of DRAGON compared to the vanilla distributed RAG vs. varying $c^l_\text{dec}$, $c^r_\text{dec}$, $\text{rtt}$ and $\alpha^r_t$.}
    \vspace{-0.2in}
    \label{fig:theoretical_speedup}
\end{figure}

Figure~\ref{fig:theoretical_speedup} illustrates the theoretical speedup characterized in Theorem~\ref{theorem:improvement}. The speedup achieves its maximum when the device-side decoding latency is minimal and maintains saturated until it surpasses that of the cloud. Thereafter, the speedup decreases inversely with $c^l_\text{dec}$, gradually approaching 1 and eventually stabilizing at 1 once $c^l_\text{dec}$ exceeds $c^r_\text{dec}+\text{rtt}$. Finally, we have following corollaries:

\begin{corollary}
    DRAGON is particularly effective when the decoding latency gap between the device and the cloud is small and the transmission cost becomes the primary bottleneck.
\end{corollary}

\noindent This property broadens the potential application of DRAGON to general scenarios in which distributed computing nodes have comparable computational resources, but communication remains a key bottleneck requiring further optimization.

\begin{corollary}
    DRAGON's improvement in wall time can be substantially amplified when the cloud-side acceptance rate is high.
\end{corollary}

\noindent Numerous existing works~\cite{kvquant,h2o} have shown that a small subset of tokens receives the majority of attention and replacing them significantly changes the output sequence~\cite{lin2024critical}. Accordingly, we argue that draft tokens that differ from those on the other side primarily originate from this subset and are synchronized across both sides. In contrast, other tokens (such as stop words, punctuations and common-knowledge terms) are often context-independent and shared across both sides, leading to a considerable acceptance rate.

\begin{corollary}
    DRAGON's improvement in wall time is independent of the device-side acceptance rate.
\end{corollary}

\noindent When the device-side decoding latency is much lower, the Aggregator must wait for the arrival of cloud-side draft tokens before generating the next target token, regardless of whether the device-side draft is accepted. Similarly, when the device-side latency is substantially higher, the next target token is generated immediately and fed as input for the next decoding step after completing the current one.
As a result, the acceptance of the local draft has no impact on the overall latency. However, it remains important when aggregation is shifted to the other side via DRAGON's scheduling algorithm.

\section{Experiments}

\subsection{Implementation}
We implemented DRAGON for distributed RAG workflow comprising \textasciitilde3,000 lines of Python code.\footnote{Our code is available at GitHub: \url{https://github.com/ThomasAtlantis/DRAGON}} The System  consists of two symmetric processes, the device-side and cloud-side ones, each utilizing eight threads for core functionalities (e.g., decoding, aggregation and transmission) along with a memory-resident service process for document retrieval. We implemented information synchronization between threads using multi-producer, multi-consumer queues, and between processes using socket-based communication. 
We utilized \textit{PyTorch}~\cite{pytorch} (version 2.6.0) for algorithm implementations, \textit{Hugging Face Transformers}~\cite{wolf2019huggingface} for LLM utilities, \textit{LangChain}~\cite{langchain} for document chunking, \textit{Sentence Transformers}~\cite{sentence-transformer} for document re-ranking, and \textit{Faiss}~\cite{faiss_gpu} for indexing and similarity search of document embeddings.

\noindent\textbf{Efficient transmission.} We implemented data transmission over the TCP/IP protocol using the \textit{socket} library. A fixed-length message header is defined using the \textit{struct} module, containing the message type and body size. All Python objects in the message body are finally serialized using the \textit{pickle.dumps()} function and compressed by means of an \textit{LZ4} compressor, while numeric vectors are first serialized with \textit{numpy.tobytes()}. For transmitting the output distributions $\bm p^\text{device}_t$ and $\bm p^\text{cloud}_t$, we employ an aggressive top-$p$ selection strategy~\cite{topp_sampling} with $p=0.8$, encoding the selected indices as unsigned 8-bit integers and the values as 16-bit floating-point numbers. While preserving the inference performance, the transmission data size is significantly reduced---by approximately 2,363 times when given the vocabulary size of 50,272---compared to the unoptimized JSON-based implementation.

\noindent\textbf{Preemptible generation.} We implemented a hook function that raises an exception upon the occurrence of a stop event (e.g., receiving a draft token rejection message) and registered it in the forward pass of each model layer to enable layer-wise interruption of draft decoding. When the generation caller catches the exception, it rolls back the KV cache and attention masks based on the number of generated target tokens so far and feeds the latest target token as input to trigger re-computation.

\subsection{Experiment Setups}\label{sec:exp_setup}
\textbf{Testbed.} 
We evaluated our framework and baseline methods using a high-performance computer as the cloud server and a MacBook Pro as the edge device. The server is equipped with an Intel Xeon Silver 4210R CPU, 64GB of memory, and a GeForce RTX 3090 GPU, while the MacBook Pro features an Intel Core i7 CPU, 16GB of memory, and no dedicated GPU. The cloud and the device are connected via a 2.4 GHz Wi-Fi local-area network, with latency and jitter measured by sockperf as 2ms and 6ms, respectively. To simulate network jitter, we replay a predefined random latency trace by adjusting the network interface controller (NIC) latency using the traffic control tool, \textit{tc}.

\noindent\textbf{Datasets and metrics.} We evaluated the long-sequence generation performance of DRAGON on the large-scale language modeling dataset WikiText~\cite{wikitext}, which comprises over 100 million tokens extracted from verified Good and Featured articles on Wikipedia. We constructed retrieval corpora from the training sets of two different-scale versions, WikiText2 and WikiText103. During evaluation, we applied rolling windows of 1024 and 512 tokens, respectively, over their test sets, using the first 1/8 of each window as the query for retrieval and the remaining tokens for perplexity evaluation. To further assess the efficiency of our method, we measure the time to first token (TTFT) and per-token latency. In this measurement, we used the retrieval corpus and index pre-built by Facebook from a Wikipedia dump dated December 20, 2018, which contains 21 million documents. 


\noindent\textbf{Models and baselines.} We evaluated our framework using OPT-1.3B~\cite{zhang2022opt} and Qwen2.5-1.5B~\cite{qwen2}, with vocabulary sizes of 151,936 and 50,272, respectively. For language modeling and latency measurement, we adopted Contriever~\cite{izacard2021contriever} and DPR~\cite{dpr} as the retrievers, respectively. Additionally, we employed ms-marco-MiniLM-L6-v2~\cite{sentence-transformer} for document re-ranking. We compare DRAGON with four baseline methods:
\begin{itemize}[leftmargin=*,topsep=2pt]
    \item {CRCG}, centralized generation augmented with centralized retrieval from local corpus, using the context-aggregation strategy, which represents most existing RAG methods~\cite{ralm,SAIL,ActiveRAG}.
    \item {DRCG}, on-device generation augmented with documents retrieved from a distributed corpus spanning both the device and the cloud, using the context-aggregation strategy.
    \item {DRDG/TW}, distributed RAG using the output aggregation strategy and token-wise synchronization, as discussed in \S~\ref{sec:vanilla_DRAG}. The target tokens are collected and aggregated on the device side.
    \item {DRDG/SW}, distributed RAG using the output aggregation strategy and sequence-wise synchronization, i.e., one-time aggregation of the independently generated output sequences from the device and the cloud. This baseline is implemented by extending the official REPLUG~\cite{replug} implementation and Facebook's RAG-Sequence model~\cite{FacebookRAG} with distributed support.
\end{itemize}
To simulate insufficient but complementary corpus in the cloud and device sides, we constrain the on-cloud and on-device retrieval by selecting the first and second halves of the top-k documents from the same corpus, respectively. Moreover, to study the overhead of DRCG, we evaluate two variants: {DRCG/Text} retrieves raw text and prefill KV cache from scratch and {DRCG/KV} retrieves and reuses the KV cache of documents directly.

\subsection{Overall Performance and Efficiency\label{sec:evaluation:overall}}
We first present the overall performance and efficiency of DRAGON in comparison to the baselines. In the following experiments, we set the maximum context length to 256 tokens on both the device and cloud sides, with each retrieved document limited to 64 tokens.

\begin{figure}[t!]
    \centering
    \includegraphics[width=\linewidth]{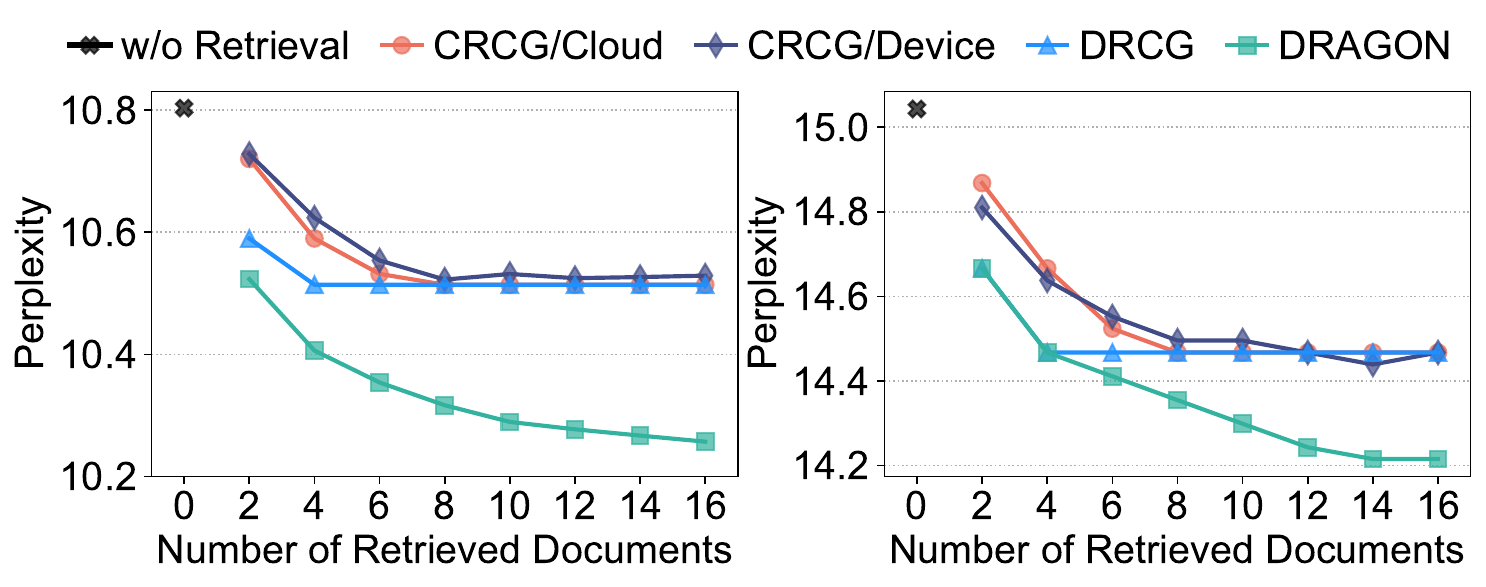}
    \vspace{-0.2in}
    \newline
    \hspace*{0.5em}
        \begin{subfigure}{0.47\linewidth}
            \caption{Qwen2.5-1.5B/WikiText2.}
            \label{fig:perplexity:wikitext2}
        \end{subfigure}
        \begin{subfigure}{0.47\linewidth}
            \caption{OPT-1.3B/WikiText103.}
            \label{fig:perplexity:wikitext103}
        \end{subfigure}
    \vspace{-0.1in}
    \caption{Performance on WikiText.}
    \vspace{-0.2in}
    \label{fig:perplexity}
\end{figure}

\noindent\textbf{Performance.} We linearly increase the number of retrieved documents on both the device and the cloud sides from 0 to 16 and report the corresponding language modeling perplexity on WikiText. As shown in Figure~\ref{fig:perplexity}, DRAGON matches or outperforms all baseline methods across all settings. As more documents are integrated, the performance gap between DRAGON and the baseline methods widens. Finally, DRAGON achieves $1.9\times$ and $1.4\times$ improvements over the non-RAG method, compared to the second-best RAG baselines, for Qwen and OPT, respectively. In contrast, CRCG methods perform poorly due to an insufficient number of retrieved documents, which indicates incomplete knowledge for the given context. Additionally, the performance of DRCG quickly saturates once the amount of retrieved text reaches the context budget limit. However, we observe a gap between DRCG and our method prior to the saturation, suggesting that output aggregation may inherently outperform context aggregation. The results of DRDG methods are omitted, as they produce identical outputs to DRAGON under the language modeling setting.

\begin{figure}[t]
    \centering
    \includegraphics[width=\linewidth]{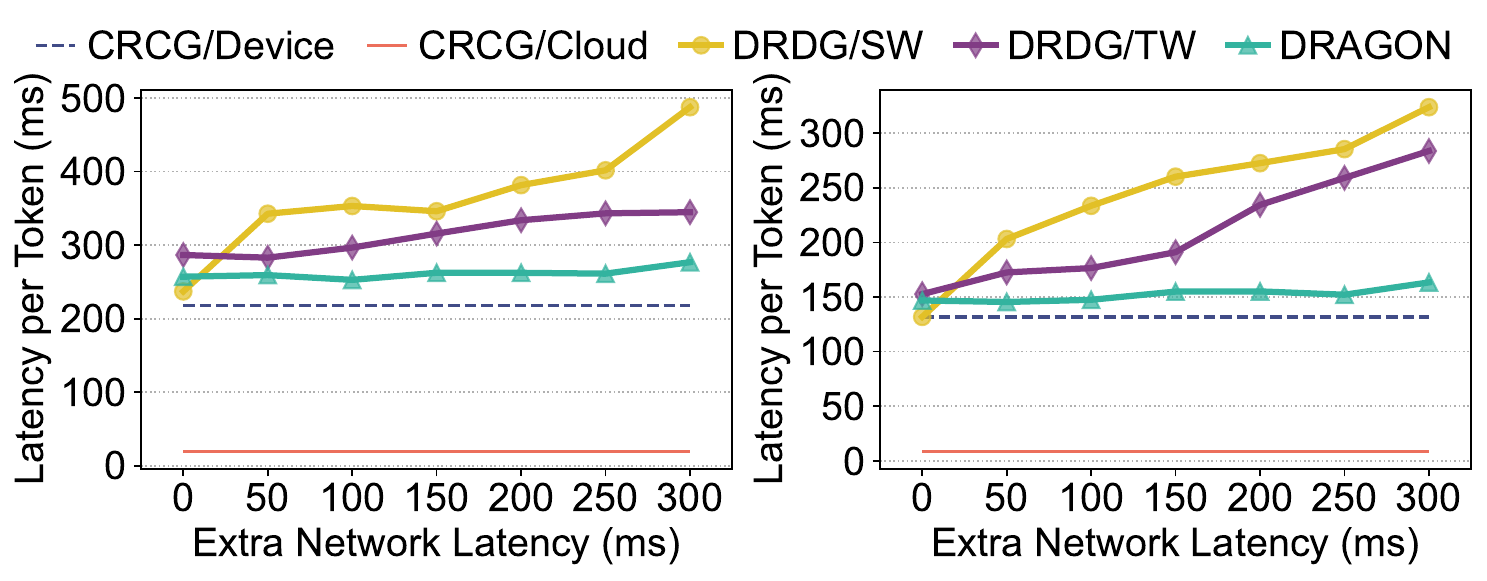}
    \vspace{-0.2in}
    \newline
    \hspace*{0.5em}
        \begin{subfigure}{0.47\linewidth}
            \caption{Qwen2.5-1.5B.}
            \label{fig:latency:qwen}
        \end{subfigure}
        \begin{subfigure}{0.47\linewidth}
            \caption{OPT-1.3B.}
            \label{fig:latency:opt}
        \end{subfigure}
    \vspace{-0.1in}
    \Description{}
    \caption{Per-token latency in various network conditions.}
    \vspace{-0.2in}
    \label{fig:latency}
\end{figure}

\begin{figure}[t]
    \centering
    \includegraphics[width=\linewidth]{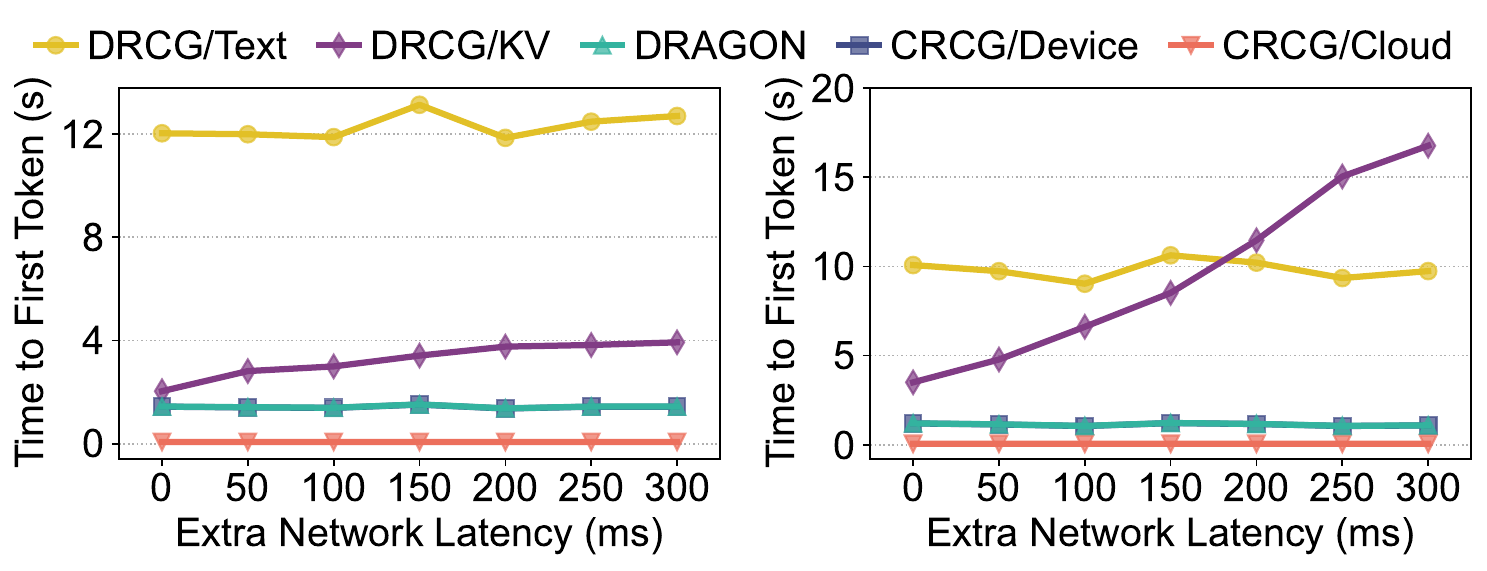}
    \vspace{-0.2in}
    \newline
    \hspace*{0.5em}
        \begin{subfigure}{0.47\linewidth}
            \caption{Qwen2.5-1.5B.}
            \label{fig:ttft:qwen}
        \end{subfigure}
        \begin{subfigure}{0.47\linewidth}
            \caption{OPT-1.3B.}
            \label{fig:ttft:opt}
        \end{subfigure}
    \vspace{-0.1in}
    \Description{}
    \caption{Time-to-First-Token in various network conditions.}
    \vspace{-0.2in}
    \label{fig:ttft}
\end{figure}

\noindent\textbf{Efficiency.} We inject additional latency to the server's NIC, ranging from 0 to 300 ms, along with a jitter equal to 1/5 of the corresponding latency value. We sample prompts from 
\textit{10k\_prompts\_ranked}~\cite{10k_prompts}, a collection of synthetic and human-generated prompts with associated ranking, and report the average end-to-end decoding latency over 20 output tokens\footnote{Despite averaging, the results still exhibits fluctuations due to varying CPU load and network jitter, but do not affect the overall conclusion.}. Figure~\ref{fig:latency} presents the per-token latency when incorporating the top-2 relevant documents for the RAG process on each side. As shown in the figure, DRAGON demonstrates strong robustness under different network conditions compared to other distributed baseline methods. Specifically, DRAGON achieves latency reduction of 49.5\% and 42.4\% when using OPT-1.3B compared to the sequence-wise and token-wise DRDG methods, respectively. In contrast, the per-token latency of DRDG methods fluctuates significantly and tends to increase under higher network latency conditions. Sequence-wise DRDG collects output distributions of all tokens once after generation completes, resulting in a one-time large data transmission and increased sensitivity to network latency. Token-wise DRDG amortizes data transmission over the entire generation process, partially hiding latency within decoding. However, it still under-performs compared to DRAGON due to frequent output synchronizations. Additionally, DRCG methods yields the same per-token latency with corresponding CRCG methods, because they do not involve cooperation between the device and the cloud. Although DRAGON incurs an average latency overhead of 15.6\%–20.3\% compared to device-only methods, it effectively supports tasks that require both personal and general knowledge, where device-only or cloud-only methods may fail.

We further compare the TTFT of DRAGON with that of the baseline methods under identical network conditions. TTFT typically includes the time for document retrieval and the latency of the prefill stage, during which the key-value (KV) activations for the concatenation of retrieved documents and the input query are either computed from scratch in parallel or loaded from cache. As shown in Figure~\ref{fig:ttft}, DRAGON incurs negligible TTFT overhead compared to the device-only CRCG method. In contrast, as KV cache is hosted on the same side with the corpus, DRCG/Text performs prefill from scratch, resulting in high computation latency and $8.6\times$ TTFT on average compared to DRAGON. DRCG/KV directly fetches KV activations from the server, leading to increased transmission time under higher network latency and yielding over $15.3\times$ TTFT compared to DRAGON, rendering it entirely impractical. Notably, DRCG/Text incurs larger prefill latency when using Qwen2.5-1.5B compared to OPT-1.3B, due to its larger number of parameters. In contrast, DRCG/KV exhibits higher TTFT on OPT-1.3B, as Qwen2.5-1.5B employs Grouped-Query Attention (GQA~\cite{GQA}) to reduce the size of KV activations. The transmission data size in DRCG/KV is 114 MB for OPT-1.3B and 16 MB for Qwen2.5-1.5B when retrieving 2 documents of 64 tokens each. Latency for local document retrieval is measured at 52.6 ms, while latency for remote raw-text retrieval ranges from 107.2 ms to 745.2 ms as extra network latency increases from 0 to 300 ms.

\begin{figure}[t]
    \centering
    \includegraphics[width=\linewidth]{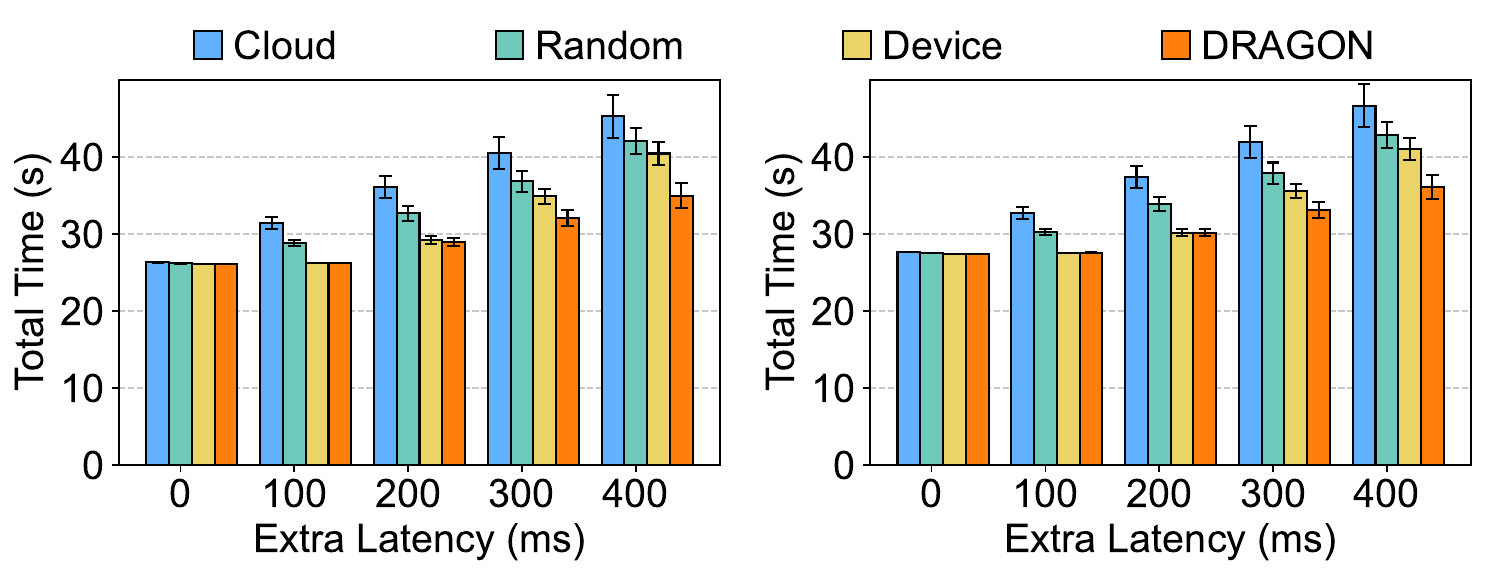}
    \vspace{-0.2in}
    \newline
    \hspace*{0.5em}
        \begin{subfigure}{0.47\linewidth}
            \caption{Qwen2.5-1.5B.}
            \label{fig:scheduling:qwen}
        \end{subfigure}
        \begin{subfigure}{0.47\linewidth}
            \caption{OPT-1.3B.}
            \label{fig:scheduling:opt}
        \end{subfigure}
    \vspace{-0.1in}
    \Description{}
    \caption{Comparison of different scheduling strategies.}
    \vspace{-5mm}
    \label{fig:scheduling}
\end{figure}

\subsection{Effectiveness of Scheduling}
To thoroughly evaluate the effectiveness of scheduling, we implemented a simulator to run DRAGON repeatedly using different scheduling strategies under consistent settings. We compare our scheduling strategy with three baseline methods: (1) \textit{Cloud} and (2) \textit{Device}, where aggregation is statically performed in the cloud and the device, respectively, and (3) \textit{Random}, which randomly selects the side for aggregation.

To implement the simulation, we record and replay the acceptance decisions of the Aggregator, and use real-world measurements of decoding latency on each side. We simulate varying network conditions by adding an extra latency and a sinusoidal jitter to the measured base latency. The period of the jitter is set to $20\pi$ seconds with its amplitude set to 1/5 of the corresponding latency, consistent with the settings in \S~\ref{sec:evaluation:overall}.

Figure~\ref{fig:scheduling} presents the total time required to generate 100 tokens under varying network conditions, each averaged over 50 different acceptance decision sequences. The results show that DRAGON's scheduling strategy matches or outperforms all baselines across all settings, with the efficiency gains increasing as the extra latency grows. Due to the substantial gap in decoding latencies between the device and the cloud (as shown in Figure~\ref{fig:latency}), performing aggregation on the device naturally hides cloud-side decoding and transmission within device-side decoding. When network latency is low, \textit{Cloud} and \textit{Random} tend to incur higher latency while DRAGON consistently selects the device side for aggregation. As network latency grows and transmission becomes the bottleneck, DRAGON dynamically selects the side with higher acceptance rate to minimize transmission resulted from draft rejection. Finally,
we argue that when device-side and cloud-side decoding latencies become closer in value, the overall generation time will be more sensitive to the network latency. In that case, our scheduling strategy will achieve greater improvement compared to these baseline methods.

\begin{figure}[t]
    \centering
    \includegraphics[width=\linewidth]{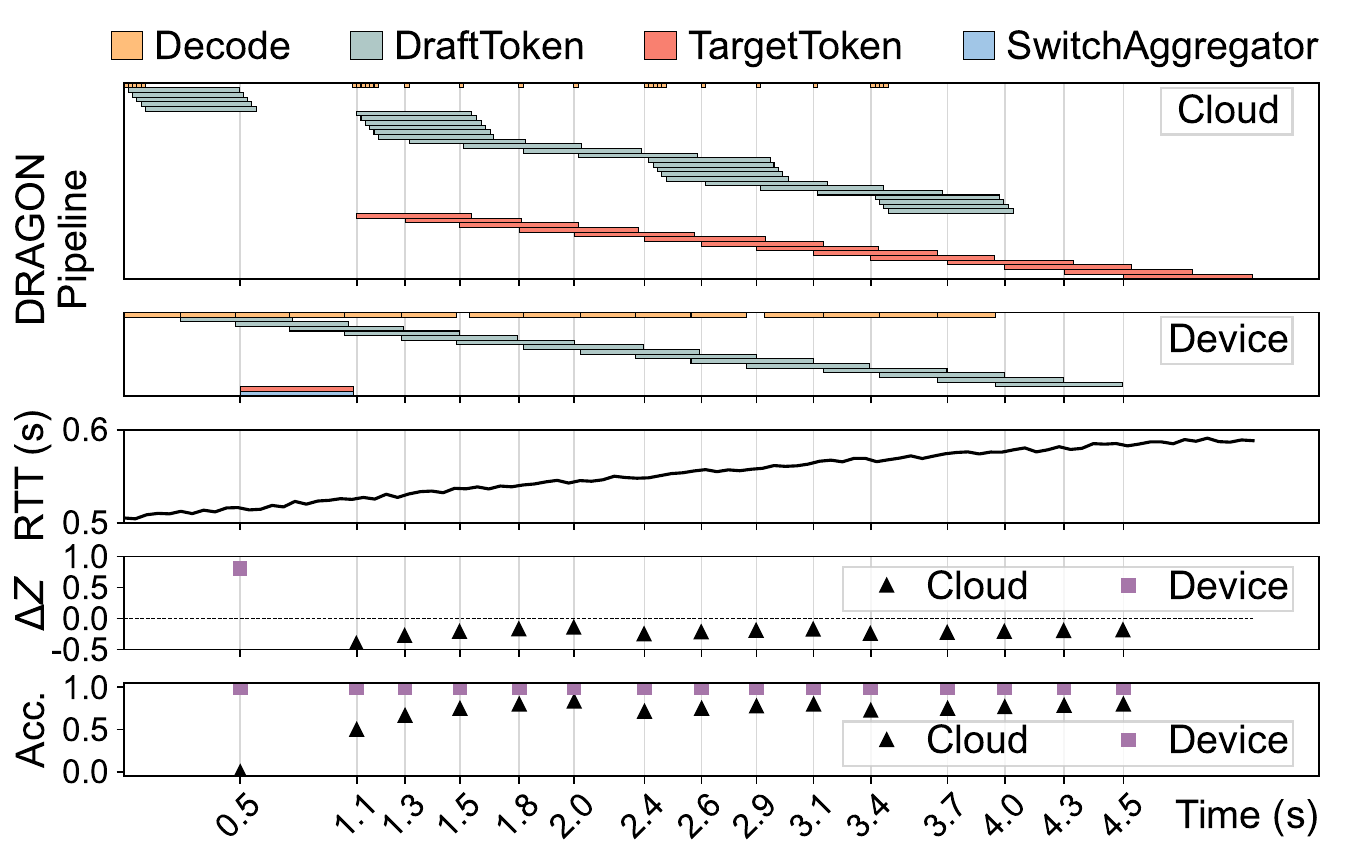}
    \vspace{-0.3in}
    \Description{}
    \caption{A random snapshot of the generation pipeline and scheduling decisions of DRAGON.}
    \vspace{-7mm}
    \label{fig:scheduling_case}
\end{figure}

\noindent\textbf{Case study.} To illustrate DRAGON's detailed scheduling process, we present a 15-token snapshot of a random simulation with the extra latency set to 500 ms. Figure~\ref{fig:scheduling_case} shows, from top to bottom, the cloud-side and device-side generation pipelines, the instantaneous RTT, the estimation score $\Delta Z$ as defined in Equation~\eqref{equa:delta_z}, and the accumulated acceptance rates. The pipeline graph comprises vertically arranged bars representing decoding and different transmission tasks (including transmission of draft tokens, target tokens and instruction signals for switching aggregation place).

Initially, the Aggregator resides on the device by default. From the perspective of the device, $c^r_\text{dec}<c^l_\text{dec}\leq c^r_\text{dec}+\text{rtt}$ consistently holds and $\Delta Z$ is computed as the sum of two terms, $A=(1-\alpha^r_t)(c^r_\text{dec}-c^l_\text{dec})$ and $B=(\alpha^l_t-\alpha^r_t)\text{rtt}$. After the first aggregation at 0.5 s, the acceptance rates are updated to $\alpha^l_0=1$ and $\alpha^r_0=0$. As a result, the positive term $B$ dominates and $\Delta Z>0$. The Scheduler decides to switch the Aggregator to the cloud, sending the switching signal along with the target token. It then shifts to the cloud's perspective and reverses the sign of $\Delta Z$. Subsequently, since the accumulated cloud-side acceptance rate remains lower, the Scheduler continues to estimating $\Delta Z<0$, indicating that cloud-side aggregation is more efficient. This case shows that DRAGON's scheduling strategy dynamically minimizes decoding and transmission costs on the side with a lower acceptance rate, which is consistent with our analysis in \S~\ref{sec:scheduling_strategy} and the results shown in Figure~\ref{fig:scheduling}.

\section{Related Works}
\textbf{RAG with multiple documents.} RAG approaches commonly retrieve multiple documents to improve performance during inference~\cite{asai2024reliable}, but the way of aggregating them primarily diverge into two categories: \textit{output aggregation} and \textit{context aggregation} (\S~\ref{sec:output_aggregation}). For output aggregation, pioneering works~\cite{FacebookRAG, realm} prior to LLMs 
have proven its effectiveness
for encoder-only and seq2seq models on both extractive~\cite{realm} and abstractive~\cite{FacebookRAG} NLP tasks. REPLUG~\cite{replug} expands this method to off-the-shelf decoder-only LLMs by fine-tuning a dense retriever.
CAD~\cite{trusting-your-evidence} adopts the same idea to strike a balance between retrieval-augmented outputs and LLM-only outputs. RA-CM3~\cite{ramlm} enables few-shot image classification for multimodal language model by aggregating the predictions given different retrieved examples. 
Context aggregation prepend the concatenation of all documents to the input and is adopted by a line of in-context RAG methods~\cite{ralm,SAIL,ActiveRAG} for simplicity. 
PCW~\cite{PCW} eliminates cross-attentions between documents to mitigate the high computational overhead introduced by this architecture. 
Our framework leverages \textit{output aggregation} to facilitate the decomposition of the multi-document RAG workflow across the device and the cloud, whereas existing works adopt a centralized architecture. 

\noindent\textbf{Device-cloud collaborative inference.} To simultaneously achieve privacy preservation and low latency in mobile computing while benefiting from the robust computational power of cloud,
numerous studies~\cite{kang2017neurosurgeon,hu2019dynamic,zhang2020towards,banitalebi2021auto,laskaridis2020spinn} have investigated device-cloud collaborative inference for conventional neural networks. Recently, this collaborative inference paradigm has been extended to large language models~\cite{llm_mec_survey,llm_mec_survey_2}. 
CE-CoLLM~\cite{jin2024collm} splits LLMs along depth and offloads deeper layers to the cloud, with a context manager to cache and reuse transmitted intermediate hidden states. Crayon~\cite{crayon} offloads difficult or non-customized tasks to a more capable cloud-hosted LLM rather than the on-device SLM. 
However, only a few existing works have explored enhancing on-device RAG with cloud-side knowledge. Hybrid-RACA~\cite{Hybrid_RACA} implements a real-time composition assistant, in which cloud-side documents are retrieved, compressed by an LLM and subsequently downloaded to enhance an on-device SLM. \cite{ding2024enhancing} utilizes user’s historical interactions with the cloud-based LLM to enhance on-device kNN-LMs~\cite{knnLLM}. These works prioritize service availability over privacy preservation, retrieving information from a single database processed by LLMs instead of employing inter-model collaborative generation. In contrast, DRAGON adopts a symmetric architecture, leveraging databases on both the device and cloud sides, enabling model collaboration without compromising document privacy.

\noindent\textbf{Speculative decoding.} Speculative decoding, initially proposed in~\cite{speculative_decoding}, accelerates the sequential decoding process of LLMs through a draft-then-verify paradigm, where at each decoding step, multiple consecutive future tokens are efficiently drafted by a small LM, and then verified in parallel by the target LLM. Concurrent studies by \cite{spec_sampling} and \cite{chen2023accelerating} introduced Speculative Sampling, extending this paradigm to support diverse sampling strategies. 
These works utilize readily available smaller language models from the same model family as the target LLM for drafting, thus avoiding additional training. 
Another line of research directly utilizes the target LLM for drafting. Medusa~\cite{medusa} and Blockwise Decoding~\cite{blockwise_decoding} integrate feed-forward network (FFN) heads into the Transformer decoder, enabling parallel generation of draft tokens per step. Other works~\cite{YangLCP024,Zhang00S0CM24,speed} have investigated early exiting and layer skipping within the target LLM to implement drafting. In contrast to speculative decoding, where a single drafter fast predicts the output of the target LLM, speculative aggregation in DRAGON verifies the consistency between outputs generated by two distinct LLMs.

\section{Conclusion}
To address privacy risks of cloud LLMs and limited capabilities of on-device SLMs, we propose DRAGON, a distributed RAG framework that enhances on-device SLMs using both personal and general knowledge without raw document transmission. DRAGON partitions the RAG workflow across device and cloud, using \textit{Speculative Aggregation} to minimize output synchronization overhead. Experimental results show that DRAGON notably improves generation quality while maintaining low latency.

\bibliographystyle{ACM-Reference-Format}
\bibliography{cites}

\appendix
\section{Appendix}
\subsection{Numerical Stability\label{sec:numerical_stable}}
We leverage the log-sum-exp trick to enhance numerical stability. Specifically, after decoding a draft token on each side $s$, the corrected value of $h^s_t$ is computed as
\begin{equation*}
    \tilde h^s_t=\log \sum\nolimits_{d\in D^s} \exp(\mathcal R(d,x_{<t})).
\end{equation*}
and is synchronized across both sides along with $\log \bm p^s_t$. During aggregation, we compute $\log\eta^s_t$ as follows:
\begin{equation*}
\begin{aligned}
    &\log \text{softmax}([\tilde{h}^l_t, \tilde h^r_t])
    =\log \frac{\exp \tilde h^s_t}{\exp \tilde h^l_t + \exp \tilde h^r_t} \\
    &=\log \frac{\sum_{d\in D^s}\exp \mathcal{R}(d,x_{<t})}{\sum_{d'\in D^l\cup D^r}\exp \mathcal{R}(d',x_{<t})}
    = \log \frac{h^s_t}{h^l_t+h^r_t}= \log(\eta^s_t).
\end{aligned}
\end{equation*}
The log of the target distribution $\log\bm p_t$ is then obtained by:
\begin{equation*}
\begin{aligned}
    &\log \sum\nolimits_{s\in\{l,r\}} \exp(\log\bm p^s_t+\log \eta^s_t)=\log (p^l_t\eta^l_t + p^r_t\eta^r_t) = \log \bm p_t.
\end{aligned}
\end{equation*}
On one hand, both the log-sum-exp and log-softmax operations are inherently numerically stable. On the other hand, since our data compression algorithm only transmits the top-$p$ values of the locally-aggregated output distributions, it effectively avoids numerical underflow of $\log \bm p^s_t$.

\subsection{Correctness of the Aggregation Strategy\label{sec:proof_1}}
We will show that for any locally-aggregated distributions $\bm p^{l}_t$ and $\bm p^{r}_t$, the target token $x_t$ produced by the aggregation strategy follows a distribution identical to that sampled from $\bm p_t=\eta^{l}_t\bm p^{l}_t+\eta^{r}_t\bm p^{r}_t$, where $\{l, r\}=\{device, cloud\}$.

First, we demonstrate that the intermediate outputs $\tilde{x}^l_t$ and $\tilde{x}^r_t$ from the two independent speculative sampling processes are indeed drawn from $\bm p_t$. Note that, since $\eta^s_t=h^s_{t}/(h^l_{t}+h^r_{t})$ for $s\in \{l,r\}$, we have $\eta^l_t+\eta^r_t =1$.

For side $l$, the probability to reject a draft token is
\begin{equation*}
    \begin{aligned}
        P(rejected)&=E_{x\sim \bm p^l_t(x)} (1- \min(1, \eta^l_t+\eta^r_t \bm p^r_t(x)/\bm p^l_t(x)))\\
        &=1-\sum \min(\bm p^l_t(x), \eta ^l_t\bm p^l_t(x)+\eta^r_t\bm p^r_t(x))\\
        &=1-\sum(\bm p^l_t(x)+\min(0,\eta^r_t(\bm p^r_t(x)-\bm p^l_t(x))))\\
        &=\eta^r_t\sum -\min(0,\bm p^r_t(x)-\bm p^l_t(x))\\
        &=\eta^r_t\sum (\bm p^l_t(x)-\min(\bm p^l_t(x),\bm p^r_t(x))).
    \end{aligned}
\end{equation*}
The adjusted distribution, from which we sample after the draft token is rejected, can be expressed as
\begin{equation*}
    \begin{aligned}
        \tilde{\bm p}_t^l(x)&=\text{norm}(\max(0, \bm p^r_t(x) - p^l_t(x))\\
        &=\text{norm}(\bm p^r_t(x)-\min(\bm p^l_t(x), \bm p^r_t(x)))\\
        &=\frac{\bm p^r_t(x)-\min(\bm p^l_t(x), \bm p^r_t(x))}{\sum_{x'} (\bm p^r_t(x')-\min(\bm p^l_t(x'), \bm p^r_t(x')))}.
    \end{aligned}
\end{equation*}
Since $\sum (\bm p^l_t(x)-\min(\bm p^l_t(x),\bm p^r_t(x)))$ is equivalent to $\sum (\bm p^r_t(x)\linebreak-\min(\bm p^l_t(x),\bm p^r_t(x)))$, $P(rejected, x=\tilde x^l_t)$, the probability that $\tilde x^l_t$ is re-sampled after rejecting $x^l_t$, is
\begin{equation*}
    \begin{aligned}
        P(rejected)\tilde{\bm p}^l_t(\tilde x^l_t)
        =\eta^r_t(\bm p^r_t(\tilde x^l_t)-\min(\bm p^l_t(\tilde x^l_t), \bm p^r_t(\tilde x^l_t))).
    \end{aligned}
\end{equation*}
Finally, the sampled token $\tilde x^l_t$ follows the distribution
\begin{equation*}
    \begin{aligned}
        P(x=\tilde x^l_t) &= P(accepted, x=\tilde x^l_t)+P(rejected, x=\tilde x^l_t)\\
        &=\bm p^l_t(\tilde x^l_t)\min(1, \eta^l_t+\eta^r_t \bm p^r_t(\tilde x^l_t)/\bm p^l_t(\tilde x^l_t))\\
        & \quad + \eta^r_t(\bm p^r_t(\tilde x^l_t)-\min(\bm p^l_t(\tilde x^l_t), \bm p^r_t(\tilde x^l_t)))\\
        &= \eta^l_t\bm p^l_t(\tilde x^l_t)+\eta^r_t \min(\bm p^l_t(\tilde x^l_t), \bm p^r_t(\tilde x^l_t))\\
        & \quad + \eta^r_t\bm p^r_t(\tilde x^l_t)-\eta^r_t\min(\bm p^l_t(\tilde x^l_t), \bm p^r_t(\tilde x^l_t)))\\
        & = \eta^l_t\bm p^l_t(\tilde x^l_t)+\eta^r_t\bm p^r_t(\tilde x^l_t)=\bm p_t(\tilde x^l_t).
    \end{aligned}
\end{equation*}
As a result, $\tilde x^l_t$ is distributed identically to tokens sampled from $\bm p_t$. Since the correctness proof for the other side $r$ is symmetric, we can conclude straightforwardly that $\tilde{x}^r_t \sim \bm p_t$.

Finally, the aggregation strategy randomly select either $\tilde x^l_t$ or $\tilde x^r_t$ as the target token $x_t$, with a uniform probability. Obviously, $x_t\sim 0.5 \bm p_t+0.5\bm p_t=\bm p_t$.

\subsection{Decoding Pipelines\label{sec:pipeline}}
Apart from the theoretical analysis of latency per token in Section~\ref{sec:scheduling}, we use pipeline graphs to illustrate scenarios where each acceptance case repeats continuously. This is not necessarily how pipelines occur in practice, but it provides us with an heuristics of the scheduling strategy. In the following discussion, we define $l$ as the side responsible for aggregation (i.e., the local side) and $r$ as the other side (i.e., the remote side). We set random delays to analyze specific cases where all time values are expressed in the same unit.

\begin{figure}[t]
    \centering
    \includegraphics[width=\linewidth]{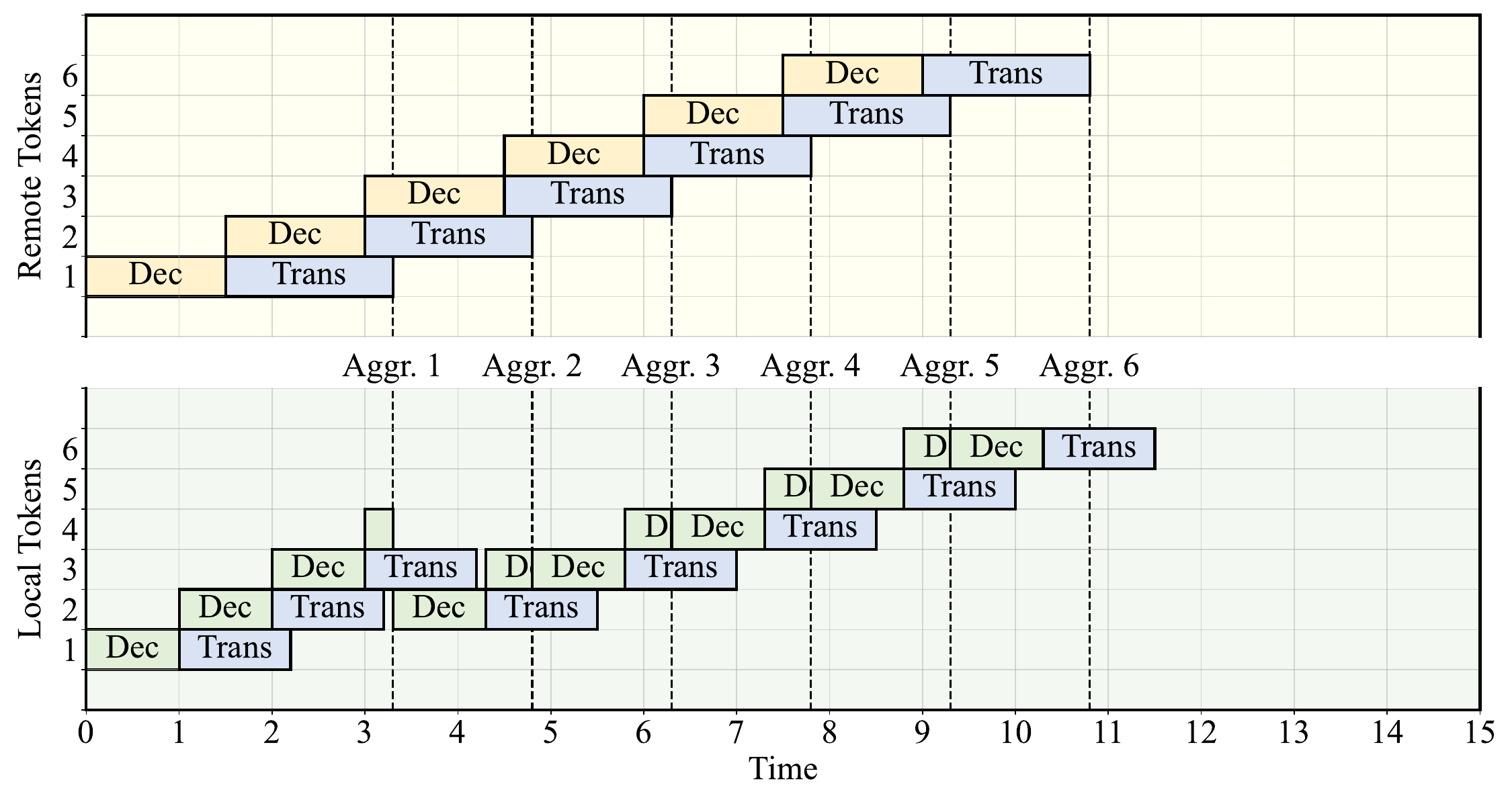}\vspace{-5mm}
    \Description{}
    \caption{Decoding pipeline when the Aggregator continuously rejects $x^l$ and accepts $x^r$.}
    \label{fig:rej_acc}
\end{figure}

\textbf{i) Continuously reject $x^l$ and accept $x^r$.} The pipeline is shown in Figure~\ref{fig:rej_acc}, where $c^l_\text{trans}=1.2$, $c^r_\text{trans}=1.8$, $c^l_\text{dec}=1$, and $c^r_\text{dec}=1.5$. The latency bottleneck is $c^r_\text{dec}$, causing $l$ to wait for $x^r_1$ before the first Aggregation. As $l$ generates draft tokens faster, subsequent aggregations begin upon the arrival of each $x^r_t$. When $r$ decodes faster, the bottleneck becomes $c^l_\text{dec}$. As a result, the latency per token is $\max(c^l_\text{dec},c^r_\text{dec})$.

\begin{figure}[t]
    \centering
    \includegraphics[width=\linewidth]{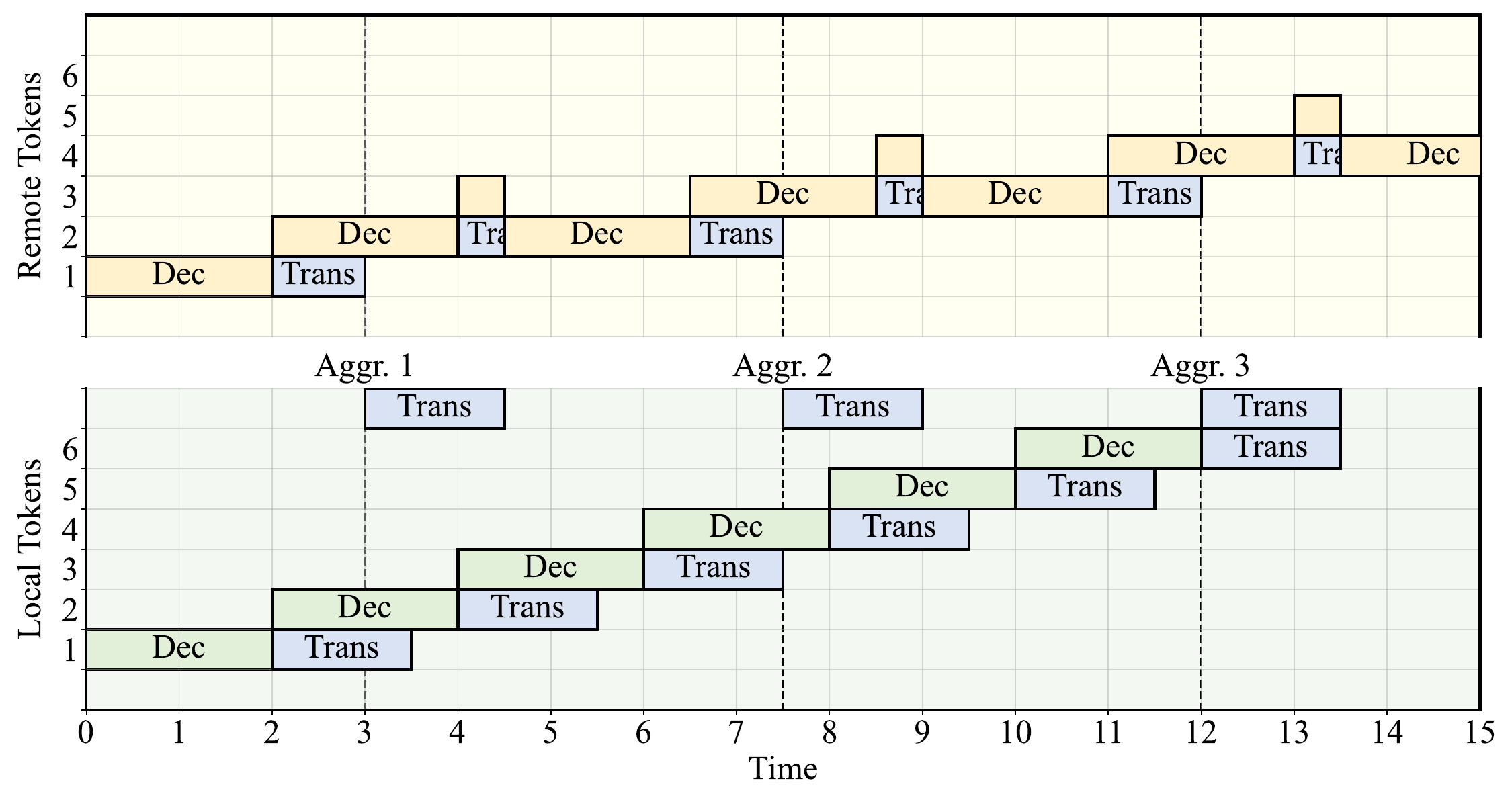}\vspace{-5mm}
    \Description{}
    \caption{Decoding pipeline when the Aggregator continuously accepts $x^l$ and rejects $x^r$.}
    \label{fig:acc_rej}
\end{figure}

\textbf{ii) Continuously accept $x^l$ and reject $x^r$.} The pipeline is shown in Figure~\ref{fig:acc_rej}, where $c^l_\text{trans}=1.5$, $c^r_\text{trans}=1$, $c^l_\text{dec}=2$, and $c^r_\text{dec}=2$. Although $c^l_\text{dec}=c^r_\text{dec}$, local draft tokens do not require transmission; therefore, the latency bottleneck thus lies on the remote side. After each aggregation at step $t$, $l$ must wait for a duration of $c^l_\text{trans}+c^r_\text{dec}+c^r_\text{trans}$, including the transmission of target token $x_{t}$, as well as the decoding and transmission of the remote draft token $x^r_{t+1}$. Only when $c^l_\text{dec}$ exceeds this duration, the bottleneck shifts to the local side. The latency per token is thus $\max(c^l_\text{dec}, c^l_\text{trans}+c^r_\text{dec}+c^r_\text{trans})$.

\begin{figure}[t]
    \centering
    \includegraphics[width=\linewidth]{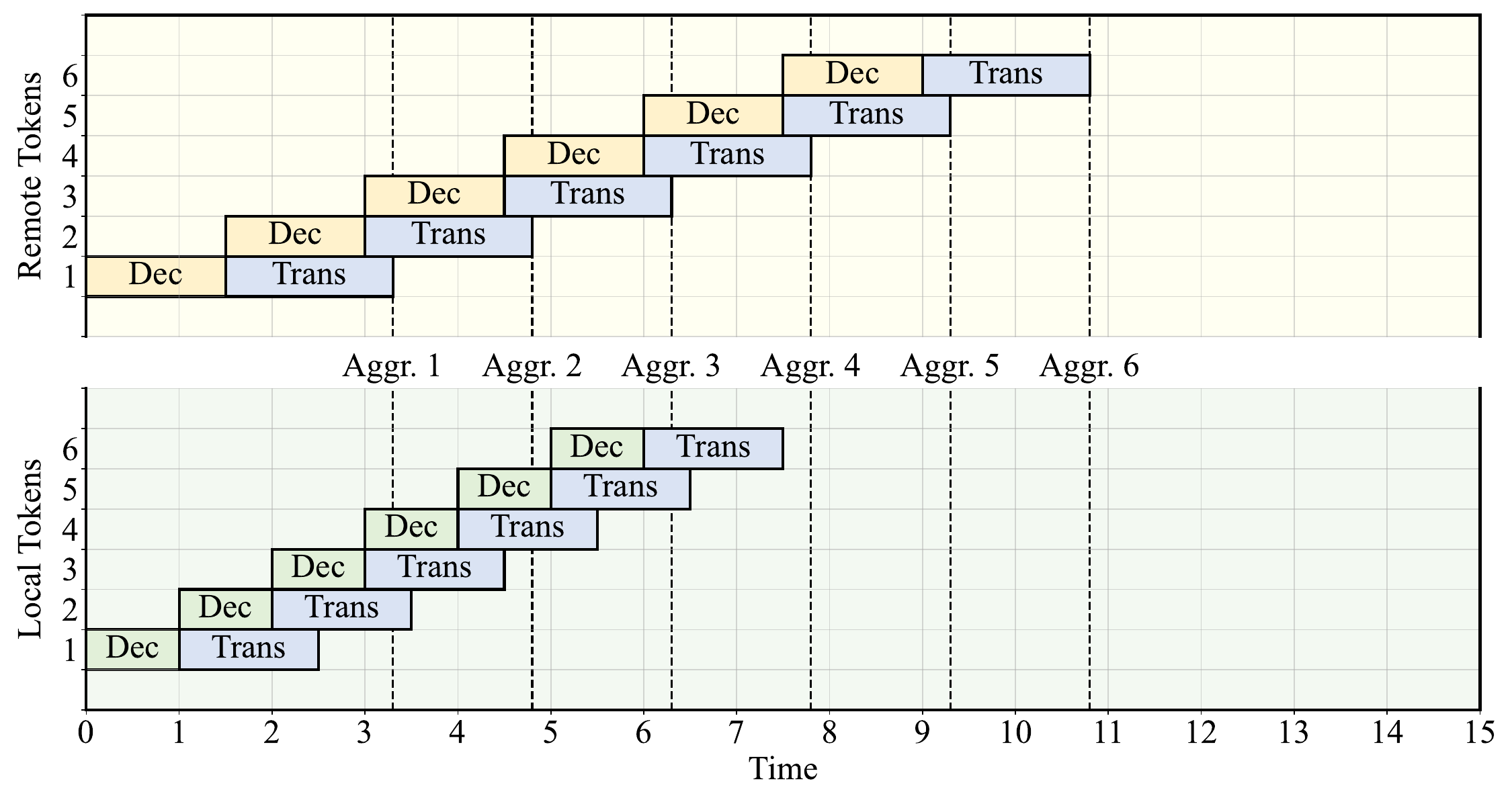}\vspace{-5mm}
    \Description{}
    \caption{Decoding pipeline when the Aggregator continuously accepts both $x^l$ and $x^r$.}
    \label{fig:acc_acc}
\end{figure}

\textbf{iii) Continuously accept both $x^l$ and $x^r$.} The pipeline is shown in Figure~\ref{fig:acc_acc}, where $c^l_\text{trans}=1.5$, $c^r_\text{trans}=1.8$, $c^l_\text{dec}=1$, and $c^r_\text{dec}=1.5$. Clearly, the bottleneck lies on the side with the larger decoding delay. Consequently, the latency per token is $\max(c^l_\text{dec}, c^r_\text{dec})$.

\begin{figure}[t]
    \centering
    \includegraphics[width=\linewidth]{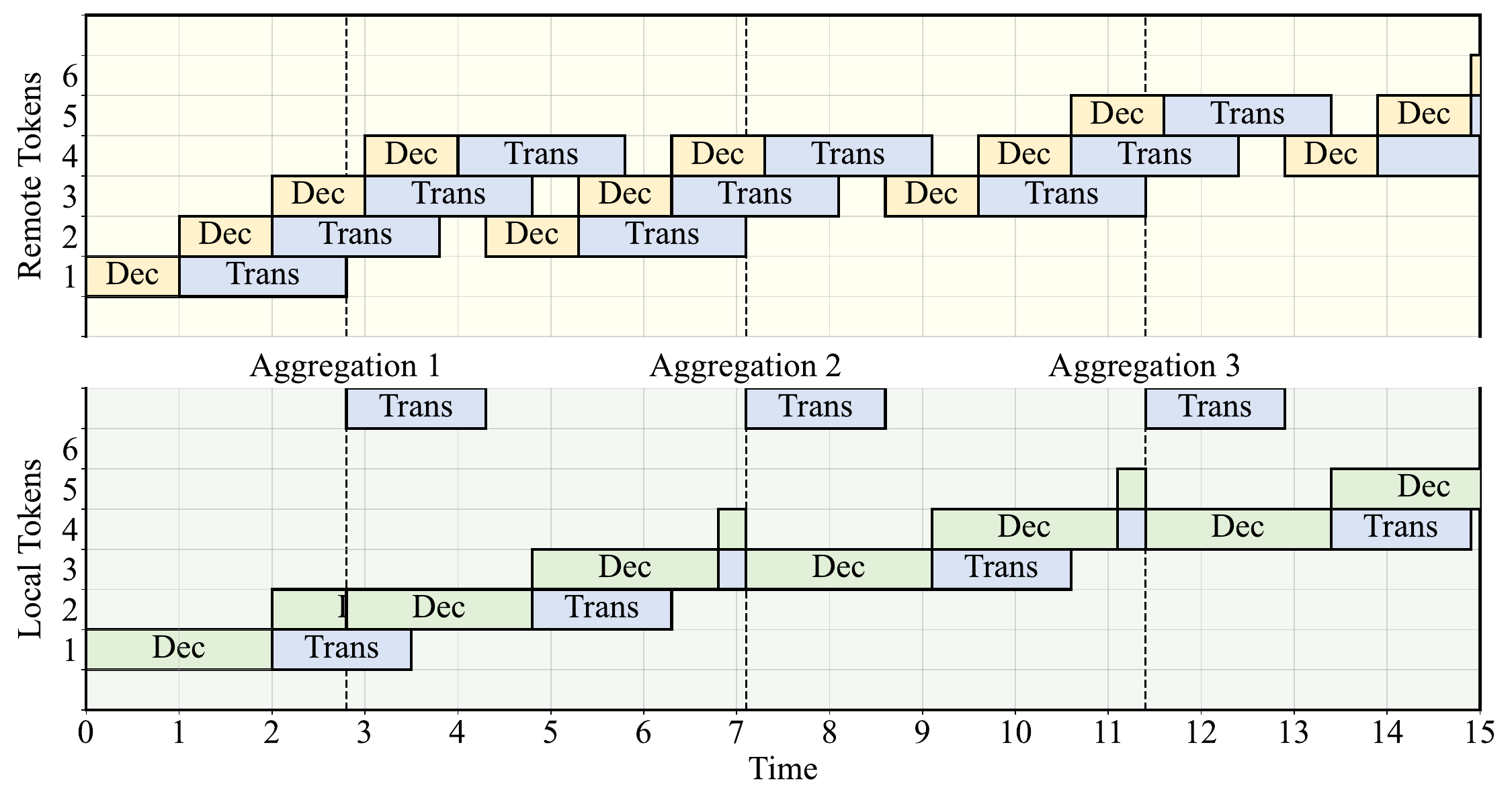}\vspace{-5mm}
    \Description{}
    \caption{Decoding pipeline when the Aggregator continuously rejects both $x^l$ and $x^r$.}
    \label{fig:rej_rej}
\end{figure}

\textbf{iv) Continuously reject both $x^l$ and $x^r$} The pipeline is shown in Figure~\ref{fig:rej_rej}, where $c^l_\text{trans}=1.5$, $c^r_\text{trans}=1.8$, $c^l_\text{dec}=2$, and $c^r_\text{dec}=1$. Since rejecting the local draft token resets $\varphi(c^l_\text{dec})$ to $c^l_\text{dec}$, the scenario is exactly the same as ii). The latency per token is computed as $\max(c^l_\text{dec}, c^l_\text{trans}+c^r_\text{dec}+c^r_\text{trans})$.

\end{document}